%% file: PaperForReview.tex
\documentclass[10pt,twocolumn,letterpaper]{article}

\usepackage[pagenumbers]{wacv} 

\usepackage{graphicx}
\usepackage{amsmath}
\usepackage{amssymb}
\usepackage{booktabs}
\usepackage{multirow}
\usepackage{makecell}
\usepackage{rotating}
\usepackage{tabularray}
\usepackage{diagbox}

\usepackage{bm}
\usepackage{adjustbox}
\usepackage{array}
\usepackage{pifont}
\usepackage{float}
\usepackage{xfrac}
\usepackage{colortbl}
\usepackage[normalem]{ulem}
\usepackage{mathtools}
\usepackage{breqn}
\usepackage{stmaryrd}

\usepackage{caption}
\usepackage{subcaption}

\usepackage{comment}
\usepackage[dvipsnames,table]{xcolor}
\usepackage{soul}
\usepackage{cite}

\newcommand\gianni{\textcolor{black}}

\newcommand\xuanlong{\textcolor{black}}
\definecolor{forestgreen}{rgb}{0.13, 0.55, 0.13}
\newcommand\antoine{\textcolor{black}}
\newcommand\david{\textcolor{black}}

\input{abbrev.tex}

\usepackage[pagebackref,breaklinks,colorlinks]{hyperref}

\usepackage[capitalize]{cleveref}
\crefname{section}{Sec.}{Secs.}
\Crefname{section}{Section}{Sections}
\Crefname{table}{Table}{Tables}
\crefname{table}{Tab.}{Tabs.}

\def\wacvPaperID{1319} 
\def\confName{WACV}
\def\confYear{2024}

\begin{document}

\title{InfraParis: A multi-modal and multi-task autonomous driving dataset}

\author{Gianni Franchi\\
U2IS, ENSTA Paris, IP Paris\\
{\tt\small gianni.franchi@ensta-paris.fr}
\and
Marwane Hariat\\
U2IS, ENSTA Paris, IP Paris\\
{\tt\small marwane.hariat@ensta-paris.fr}
\and
Xuanlong Yu\\
SATIE, Paris-Saclay University; 
U2IS, ENSTA Paris, IP Paris\\
{\tt\small xuanlong.yu@ensta-paris.fr}
\and
Nacim Belkhir\\
Safrantech, Safran Group\\
{\tt\small nacim.belkhir@safrangroup.com}
\and
Antoine Manzanera\\
U2IS, ENSTA Paris, IP Paris\\
{\tt\small antoine.manzanera@ensta-paris.fr}
\and
David Filliat\\
U2IS, ENSTA Paris, IP Paris\\
{\tt\small david.filliat@ensta-paris.fr}
}

\twocolumn[{
\renewcommand\twocolumn[1][]{#1}%
\maketitle
\begin{center}
    \centering
    \captionsetup{type=figure}
    \captionsetup[subfigure]{labelformat=empty}
     \addtocounter{figure}{-1}
     \begin{subfigure}[b]{0.26\textwidth}
         \centering
         \includegraphics[width=\textwidth]{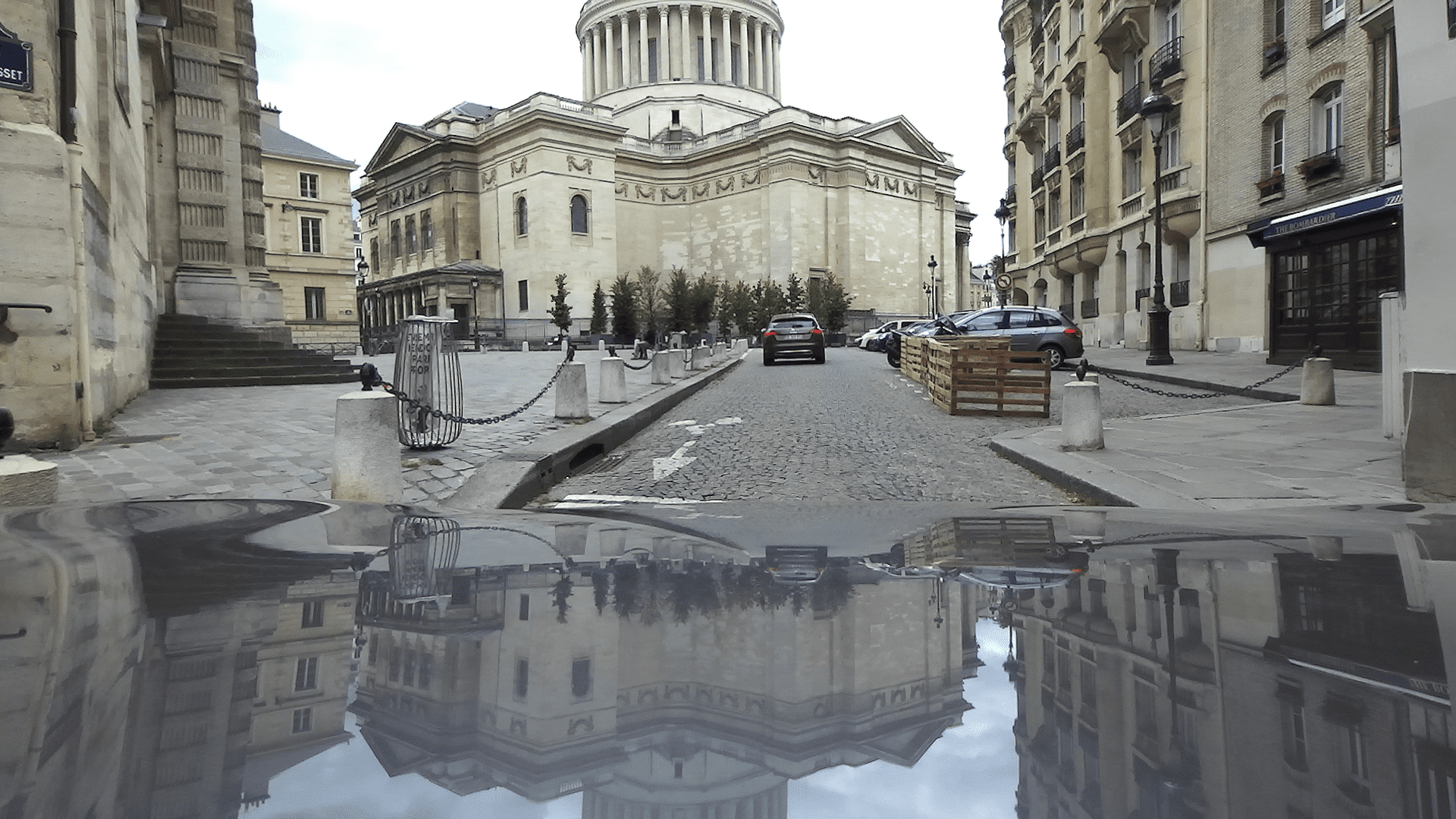}
         \label{figteaser:Label-anom}
     \end{subfigure}
     \begin{subfigure}[b]{0.22\textwidth}
         \centering
         \includegraphics[width=\textwidth]{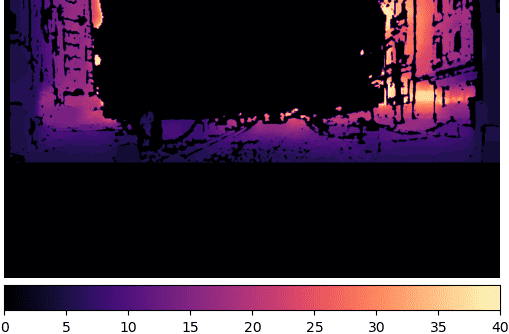}
         \label{figteaser:Orginalimage-anom}
     \end{subfigure}
     \begin{subfigure}[b]{0.17\textwidth}
         \centering
         \includegraphics[width=\textwidth]{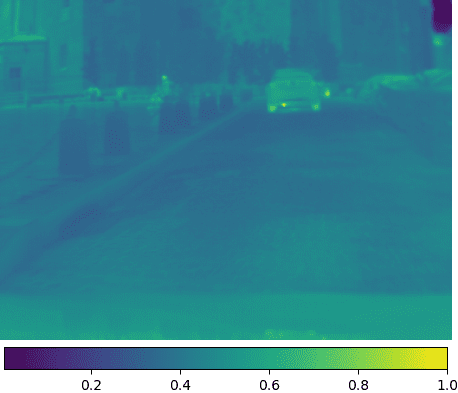}
         \label{figteaser:RobustaGeneration-anom}
     \end{subfigure}
        \begin{subfigure}[b]{0.26\textwidth}
         \centering
         \includegraphics[width=\textwidth]{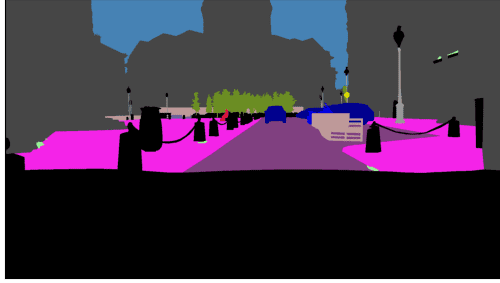}
         \label{figteaser:OasisGeneration-anom}
     \end{subfigure}
         \begin{subfigure}[b]{0.26\textwidth}
         \centering
         \includegraphics[width=\textwidth]{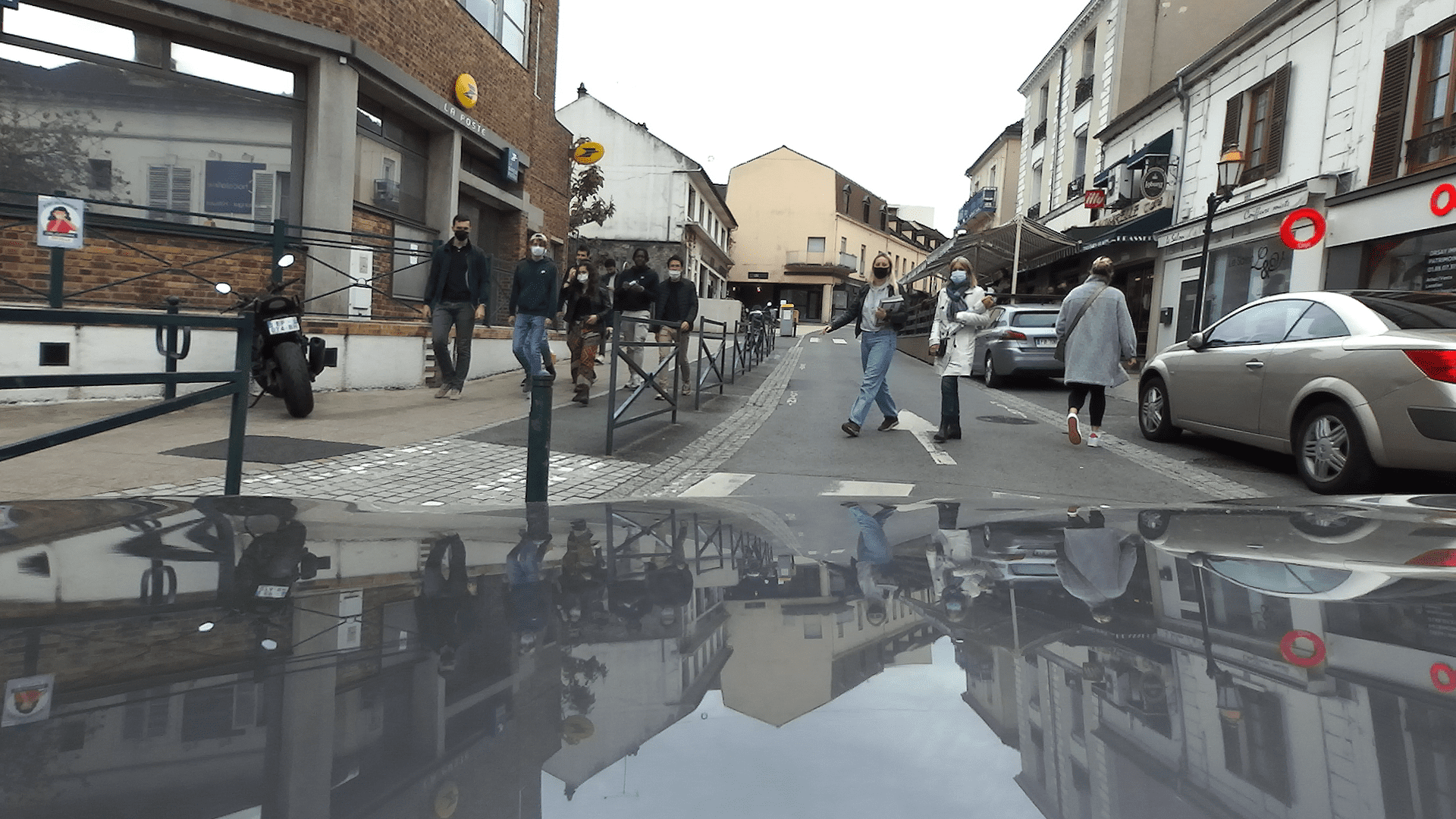}
         \caption{
         \antoine{RGB}
         images}
         \label{figteaser:Label}
     \end{subfigure}
     \begin{subfigure}[b]{0.22\textwidth}
         \centering
         \includegraphics[width=\textwidth]{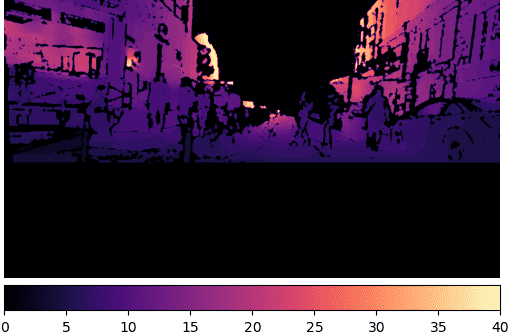}
         \caption{Depth maps}
         \label{figteaser:Orginalimage}
     \end{subfigure}
     \begin{subfigure}[b]{0.17\textwidth}
         \centering
         \includegraphics[width=\textwidth]{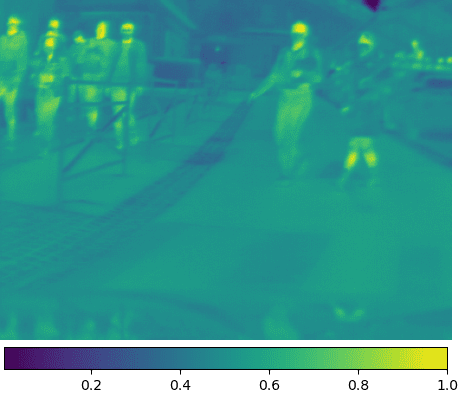}
         \caption{Infrared images}
         \label{figteaser:RobustaGeneration}
    \end{subfigure}
         \begin{subfigure}[b]{0.26\textwidth}
         \centering
         \includegraphics[width=\textwidth]{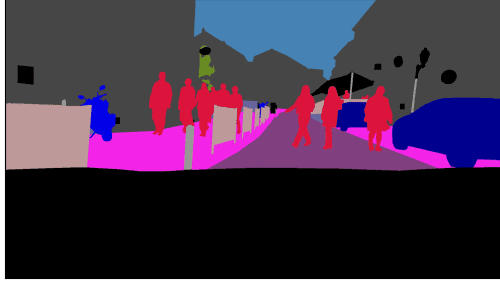}
         \caption{Segmentation maps}
         \label{figteaser:OasisGeneration}
     \end{subfigure}

    \captionof{figure}
    {\xuanlong{\textbf{Snapshots from the InfraParis dataset} showing different modalities and full semantic annotations for autonomous driving.}
    }
    \label{fig:qualitative_robustness}
\end{center}}]

\begin{abstract}
\vspace{-0.5em}
Current deep neural networks (DNNs) for autonomous driving computer vision are typically trained on specific datasets that only involve a single type of data and urban scenes. Consequently, these models struggle to handle new objects, noise, nighttime conditions, and diverse scenarios, which is essential for safety-critical applications. Despite ongoing efforts to enhance the resilience of computer vision DNNs, progress has been sluggish, partly due to the absence of benchmarks featuring multiple modalities. We introduce a novel and versatile dataset named InfraParis that supports multiple tasks across three modalities: RGB, depth, and infrared. We assess various state-of-the-art baseline techniques, encompassing models for the tasks of semantic segmentation, object detection, and depth estimation.
More visualizations and the download link for InfraParis are available at \href{https://ensta-u2is.github.io/infraParis/}{https://ensta-u2is.github.io/infraParis/}.   
\end{abstract}

\section{Introduction}
\label{sec:intro}
In the realm of autonomous driving, the ability of vehicles to navigate effectively under varying lighting conditions, including both day and night, is of paramount importance. Achieving this requires the development of intelligent systems that possess robust visual perception capabilities. In recent years, the potential of utilizing Infrared (IR) datasets to enhance vision in challenging environments has garnered attention \cite{farooq2023decisivedatausing,vadidar2022robustenvironment, kampmann2022investigatingoutdoor}. Despite this promise, the availability of comprehensive IR datasets remains limited compared to the abundance of RGB datasets \cite{Cordts2016Cityscapes, geiger2012we,  richter2016playing, ros2016synthia, tung2017raincouver}. While RGB datasets have proliferated, datasets containing IR imagery have not attained the same prevalence.

The 
\antoine{benefit}
of IR images lies in their capacity to convey thermal information, which can provide invaluable insights for understanding and interpreting the environment. However, harnessing the potential of IR datasets is not without its challenges. The distinctive nature of information captured by IR cameras introduces complexities that make dataset handling and analysis a formidable task. These intricacies are exemplified through the existence of materials 
\antoine{acting as pure mirror for the IR spectrum (e.g. water),}
thereby inducing the occurrence of false positives.
\antoine{Poor contrast and drastic changes in object appearance over time are other elements}
that hinder the popularity of IR modalities in urban scenes.

The emergence of foundation models \cite{radford2021learning,alayrac2022flamingo} capable of accommodating multimodal datasets and performing multiple tasks has marked a significant leap forward in the field. Models such as \cite{sun2023dimefmdistilling, 2021learningtransferable}
demonstrate the potential of integrating diverse data modalities to enhance overall performance. However, harnessing the benefits of these foundation models necessitates access to well-annotated multimodal datasets, a critical resource that remains scarce.

The importance of establishing a multimodal and \gianni{and multitask} dataset like the one we present, named InfraParis, becomes evident when considering challenges related to domain adaptation. The drastic dissimilarity between IR images and traditional RGB data makes the InfraParis database uniquely valuable for addressing domain adaptation scenarios. \gianni{Additionally, the incorporation of multitask learning enables DNNs to attain more generalized representations. Thus, t}his dataset can serve as an instrumental bridge for testing and improving the robustness of models across disparate modalities \gianni{and tasks.} Furthermore, the inclusion of thermal data in InfraParis opens doors for enhanced object detection, tracking, \gianni{depth estimation,} and semantic segmentation, areas where traditional RGB data might fall short.

The dataset was collected across the Parisian metropolitan area and its environs, encompassing a diverse array of scenes that exhibit varying characteristics. This geographic breadth offers a spectrum of scenes, ranging from urban settings to rural landscapes and highway roads. Consequently, the dataset encapsulates the distinct attributes associated with each of these settings, including the differing people density, crowd dynamics, and environmental conditions. Another interesting aspect is that the timing of data acquisition coincided with the extensive preparations underway for the forthcoming Olympic Games in Paris. This temporal context further enhances the dataset's complexity. The significant amount of ongoing construction and activities related to the Games introduces a high degree of variability and challenge to the dataset. The dynamic nature of this context translates into scenes featuring intricate interactions between vehicles, pedestrians, and various urban elements. This intricate tapestry of activities, coupled with the diverse surroundings, makes the InfraParis dataset uniquely demanding and reflective of real-world scenarios that autonomous vehicles might encounter.

In summary, this paper introduces the InfraParis multimodal dataset, acknowledging the necessity of IR data for comprehensive environmental perception, the challenges inherent to IR dataset management, the potential benefits of integrating multimodal foundation models, and the pivotal role of the database in tackling domain adaptation difficulties. By providing researchers with a rich resource that bridges the gap between RGB and IR domains, InfraParis paves the way for more adaptable and reliable autonomous driving systems, even in the most challenging visual conditions.


\section{Related works}

\begin{table*}[t!]
\begin{center}
\resizebox{0.85\linewidth}{!}{
\begin{tabular}{lllllllll} 
\toprule
\xuanlong{Dataset} & \xuanlong{Scenario} & Annotation type & \# images & \# classes & RGB & Depth & 
\antoine{IR}
range \\ 
\toprule
\href{https://public.roboflow.com/object-detection/thermal-dogs-and-people/1}{Thermal Dogs and People}~\cite{thermalDogsAndPeople} & Humans and dogs in infrared & Bounding box & 203 & 2 & x & x & \xuanlong{Unknown} \\ 
\midrule
\href{https://arxiv.org/pdf/1801.05944.pdf}{PTB-TIR}~\cite{liu2019ptb} & Pedestrians detection & Bounding box & 30128 & 1 & x & x & \xuanlong{Unknown}  \\ 
\midrule
\href{https://docs.google.com/forms/d/e/1FAIpQLSc7ZcohjasKVwKszhISAH7DHWi8ElounQd1oZwORkSFzrdKbg/viewform}{NVGesture}~\cite{Nvgesture} & Hand gesture recognition & \xuanlong{Hand gesture label }& 1532 & 25 & \checkmark & \checkmark & 
\antoine{0.85-0.87 \textmu m (NIR)}
\\ 
\midrule
\href{https://drive.google.com/drive/folders/1ZF2vDk9j69kP5U0zcp-liOBk-atWcw-5}{SODA}~\cite{li2020segmenting} & Image segmentation & Polygons & 7168 & 20 & x & x & 
\antoine{0.75-13 \textmu m}
\\ 
\midrule
\href{https://soonminhwang.github.io/rgbt-ped-detection/}{KAIST Multispectral Pedestrian}~\cite{pedestrianDataset}  & Pedestrians detection & Bounding box & 95000 & 3 & \checkmark & x &  
\antoine{7.5-13 \textmu m (LWIR)}
\\ 
\midrule
\href{https://github.com/unizard/kaist-allday-dataset}{KAIST all-day dataset}~\cite{choi2018kaist}  &  \begin{tabular}[c]{@{}l@{}}Autonomous driving\\in day and night\end{tabular} & Bounding box & 8970 & 3 & \checkmark & \checkmark  &  
{7.5-14 \textmu m (LWIR)}
\\ 
\midrule
\href{https://www.flir.com/oem/adas/adas-dataset-form/}{Flir thermal dataset}~\cite{FlirDataset}  & Autonomous Driving & Bounding box & 14000 & 5 & \checkmark & x & 
\antoine{7.5-13.5 \textmu m (LWIR)}
\\ 
\midrule
\href{https://github.com/Robotics-BUT/Brno-Urban-Dataset}{Brno-Urban-Dataset}~\cite{ligocki2020brno}  & Autonomous Driving & None & 13h44min &  & \checkmark & \checkmark & 
\antoine{7.5-13.5 \textmu m (LWIR)}
\\ 
\midrule
\href{http://thermal.cs.uni-freiburg.de/}{Freiburg Thermal}~\cite{vertens20bridging}                         & \xuanlong{\begin{tabular}[c]{@{}l@{}}Autonomous driving\\in day and night\end{tabular}} & Instance semantics & 20647            & 13 & \checkmark & x & 
\antoine{8-14 \textmu m (LWIR)}
\\
\midrule
\href{https://www.mi.t.u-tokyo.ac.jp/static/projects/mil_multispectral/}{MFNet dataset}~\cite{takumi2017multispectral}  & \xuanlong{\begin{tabular}[c]{@{}l@{}}Autonomous driving\\in day and night\end{tabular}} & \xuanlong{\begin{tabular}[c]{@{}l@{}}Bounding box\\Instance semantics\end{tabular}} & 1569 & 8 & \checkmark & x & 
\antoine{8-14 \textmu m (LWIR)}
\\ 
\midrule
\href{https://hal.archives-ouvertes.fr/hal-01975285/document}{\begin{tabular}[c]{@{}l@{}}All-weather vision for automotive\\safety \& all weather visibility for cars\end{tabular}}~\cite{whichSpectralBand} & \xuanlong{Autonomous driving} & \xuanlong{None} & \xuanlong{Unknown} & \xuanlong{-} & \checkmark & x &\begin{tabular}[c]{@{}l@{}}
\antoine{0.4-1.0 \textmu m (visible+NIR);}
\\
\antoine{0.6-1.7 \textmu m; 8-12 \textmu m}
\end{tabular} \\ 
\midrule
Ours & Autonomous driving & \xuanlong{\begin{tabular}[c]{@{}l@{}}Bounding box\\Full semantics\end{tabular}} & 7301 & \xuanlong{19} & \checkmark & \checkmark  & 
\antoine{8-14 \textmu m (LWIR)}
\\
\bottomrule
\end{tabular}
}
\caption{\textbf{Comparative overview} of the different IR/RGB datasets designed for different scenarios.}\label{table:IRdataset}
\vspace{-1em}
\end{center}
\end{table*}

\subsection{Autonomous driving datasets}
\label{subsec:auto_driving_dataset}
A range of real-world datasets tailored for autonomous driving purposes has recently been unveiled~\cite{Cordts2016Cityscapes, geiger2012we,waymo,nuscenes,yu2020bdd100k, varma2019idd,ramanishka2018toward,chang2019argoverse,houston2020one,franchi2022muad,shift2022}. These datasets have played a pivotal role in driving significant advancements in the field, although they typically center on singular tasks such as semantic segmentation~\cite{Cordts2016Cityscapes,yu2020bdd100k,ramanishka2018toward}, object detection~\cite{geiger2012we,waymo,nuscenes}, or motion prediction~\cite{chang2019argoverse,houston2020one}, often lacking \xuanlong{multi-task capabilities with integration of multimodal information especially infrared}. While synthetic datasets like GTA-V \cite{richter2016playing}, SYNTHIA \cite{ros2016synthia}, virtual KITTI \cite{gaidon2016virtual}, MUAD \cite{franchi2022muad}, and SHIFT~\cite{shift2022} offer ample training data without incurring the costs of annotating real images or privacy concerns, even these synthetic images fall short of including infrared data.

Other existing datasets predominantly serve domain adaptation, typically emulating content and classes from a specific real dataset. Some datasets, like Fishyscapes~\cite{blum2019fishyscapes}, Lost and Found~\cite{pinggera2016lost}, and SegmentMeIfYouCan~\cite{chan2021segmentmeifyoucan}, emphasize reliability for self-driving vehicles by evaluating semantic segmentation DNNs in the context of out-of-distribution objects. Other datasets assess robustness against varying weather conditions, including night~\cite{dai2020curriculum, dai2018dark, sakaridis2021acdc}, rain~\cite{tung2017raincouver, sakaridis2021acdc}, and fog~\cite{sakaridis2018semantic,sakaridis2021acdc}, though they often suffer from differing locations and conditions, leading to performance drops coinciding with challenging weather conditions.

To bolster the reliability of semantic segmentation DNNs and address the dearth of diversity in real-world environments, certain studies have promoted virtual object inpainting~\cite{hendrycks2019anomalyseg} or synthesis of weather conditions~\cite{tremblay2021rain}. However, these approaches raise concerns about result fidelity. The recent ACDC dataset~\cite{sakaridis2021acdc}, composed exclusively of real images from consistent locations and inclusive of aleatoric uncertainty sources, endeavors to alleviate these concerns. Yet, even with their significance, these datasets often remain confined to a single modality or task.

\subsection{\xuanlong{Multi-modal datasets and DNNs}}
\label{subsec:multi_modal_dnn}
\xuanlong{Several infrared autonomous driving datasets exist~\cite{FlirDataset,ligocki2020brno,whichSpectralBand,ha2017mfnet}, yet most lack annotations for semantic segmentation. Freiburg Thermal dataset~\cite{vertens20bridging} and MFNet dataset~\cite{takumi2017multispectral} have semantic segmentation annotations, yet the former lacks object detection annotations, and the latter is short on the image amount compared to other datasets. Moreover, the depth modality lacks in the above-mentioned datasets.
Additionally, there are some other infrared datasets~\cite{Nvgesture,thermalDogsAndPeople,choi2018kaist,liu2019ptb,pedestrianDataset} working for other scenarios such as traffic surveillance. In essence, multimodal datasets and benchmarks in autonomous driving scenarios are still under-acquired and under-explored.}

Notably, some studies have explored thermal image semantic segmentation. Qiao et al.~\cite{qiao2017thermal} employ a level set method to segment pedestrians in thermal images, while Li et al.~\cite{li2020segmenting} propose an edge-conditioned segmentation network for thermal images trained on a dataset encompassing various indoor and outdoor scenes. Ha et al.~\cite{ha2017mfnet} introduce a multimodal fusion network architecture for RGB and thermal images, evaluated on their own MF dataset~\cite{ha2017mfnet}. 

To summarize the existing landscape of Infrared datasets, we have provided an overview in Table~\ref{table:IRdataset} with a comparison to our InfraParis dataset.


\section{The dataset: InfraParis}
\label{sec:DATASET}

\subsection{Acquisition process}

\david{We used the Stereolabs ZED 2 stereo camera, capable of capturing paired color images to generate depth maps.} During the data acquisition phase, the ZED 2 played a pivotal role in achieving precise calibration and registration of depth and RGB information. We also employed the optris PI 450i Infrared camera, featuring an $80^{\circ} \times 54^{\circ}$ field of view with a spectral range of 8 – 14 $\mu$m, while the ZED 2 boasts a wider field of view at $110^{\circ} \times 70^{\circ}$, coupled with a depth range spanning \xuanlong{0 to 40} meters. Both cameras were rigidly affixed together to prevent movement and securely mounted on the vehicle's hood to mitigate potential glass distortion. The calibration process is elaborated upon in section \ref{subsec:calibration}, along with the synchronization mechanism that ensures simultaneous image capture for both cameras.

Having synchronized the database and fine-tuned camera calibration, the focus shifted to generating annotations exclusively for the RGB images, which could then be transferred to correspond with the infrared and depth counterparts. To ensure accurate and reliable annotations, we enlisted the expertise of professional annotation services. These annotations were meticulously crafted to align with the class schema established by the cityscape dataset \cite{Cordts2016Cityscapes}, thereby facilitating potential domain adaptation from cityscape to InfraParis. Rigorous quality assessment was undertaken through the collaborative efforts of university members and students using specialized annotation software. Following multiple iterations of correction and assessment, the annotations attained a commendable level of precision and fidelity.

\subsection{Ethics and policy}

We ensured participant awareness regarding their inclusion in the dataset by providing them the option to request the removal of their images through email communication within the vehicle. Following a two-year period, no such requests were received. To uphold privacy and prevent the disclosure of personal information, \gianni{we take a proactive approach by promptly removing images upon receiving complaints.}
This action underlines our commitment to respecting participants' rights and maintaining confidentiality. A notable facet of this dataset is its timeframe, captured in the aftermath of the COVID-19 pandemic. Consequently, a significant majority of individuals were photographed wearing masks. This situation significantly contributes to the heightened preservation of their privacy.

Our dataset is characterized by its comprehensive representation of diverse viewpoints and demographics. We have proactively undertaken efforts to encompass a broad spectrum of backgrounds, cultures, and life experiences. By doing so, we have diligently avoided potential biases or instances of under-representation that could distort findings or perpetuate inequalities.

\begin{figure*}[!h]
    \centering
    \begin{subfigure}[b]{0.25\textwidth}
        \includegraphics[width=\textwidth]{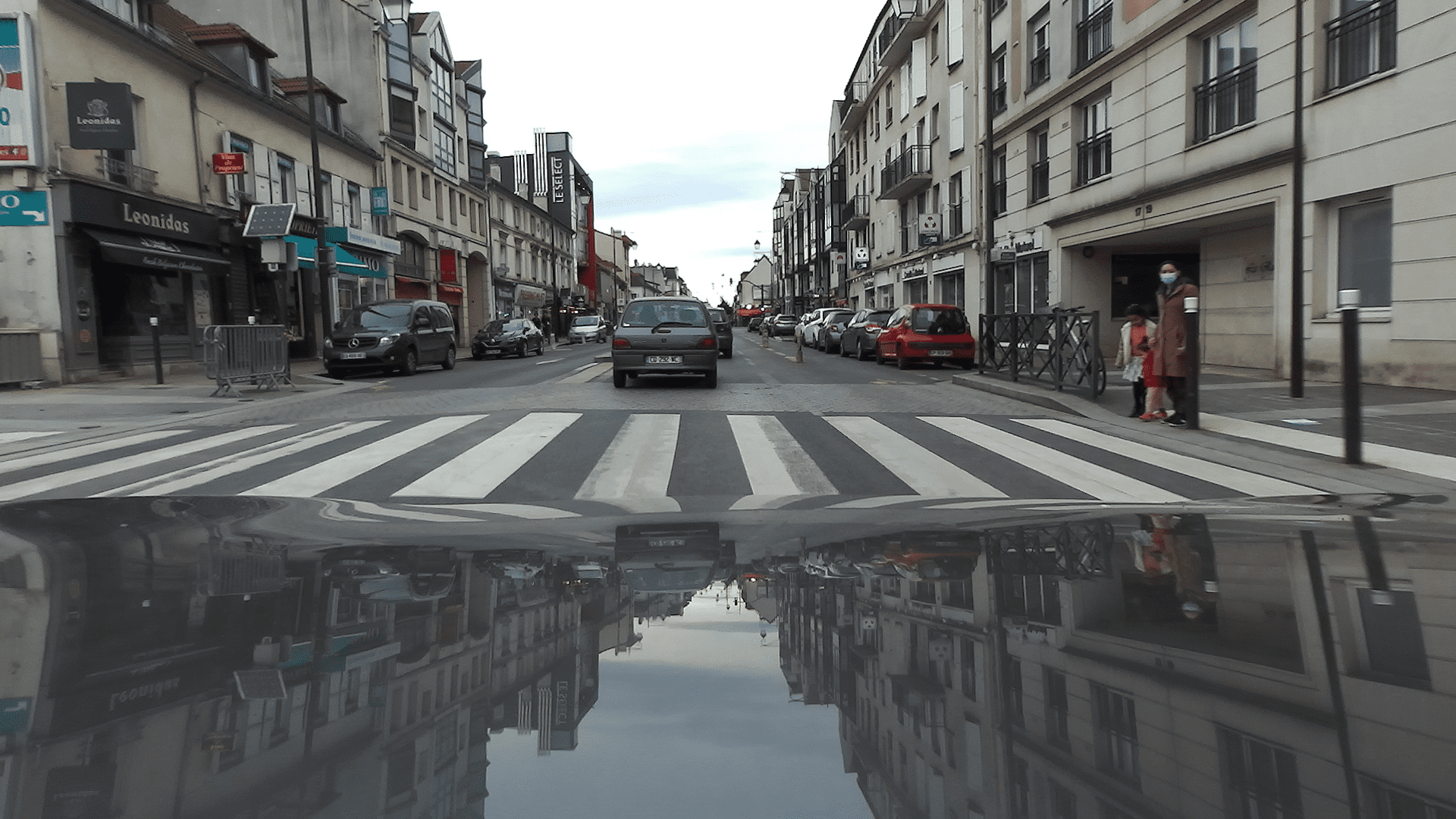}
        \includegraphics[width=\textwidth]{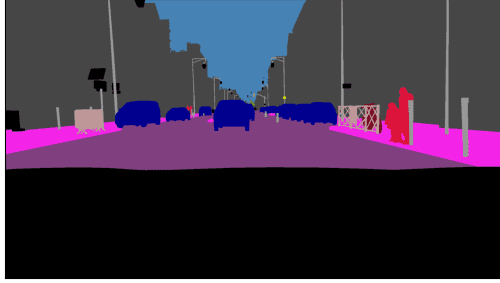}
    \end{subfigure}
    \begin{subfigure}[b]{0.25\textwidth}
        \includegraphics[width=\textwidth]{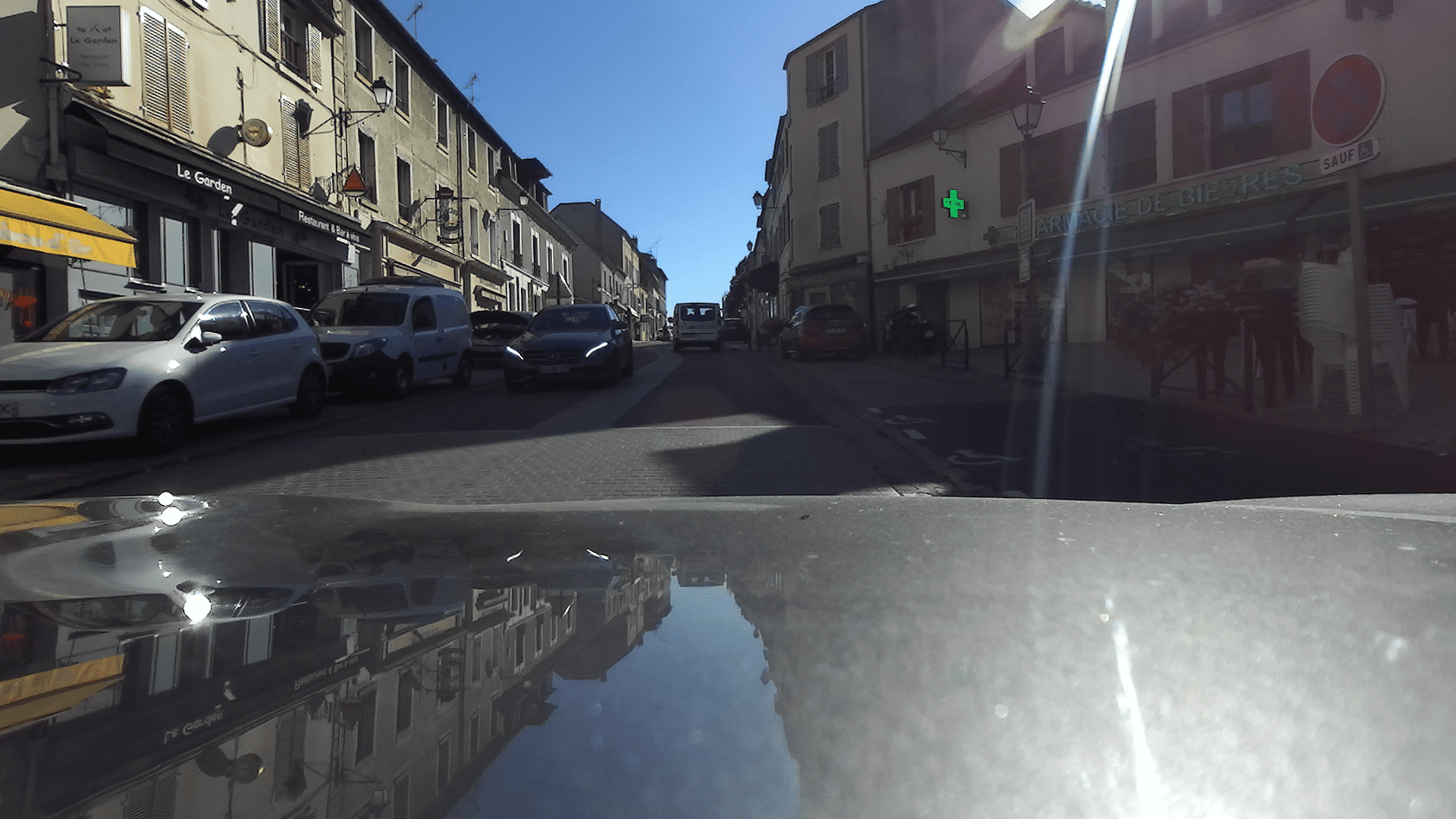}
        \includegraphics[width=\textwidth]{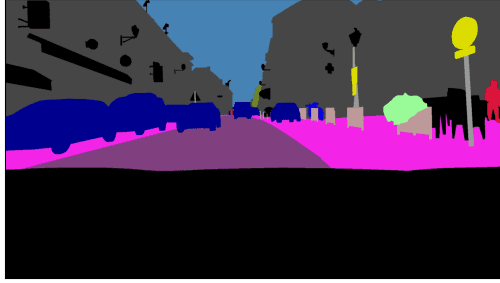}
    \end{subfigure}
    \begin{subfigure}[b]{0.25\textwidth}
        \includegraphics[width=\textwidth]{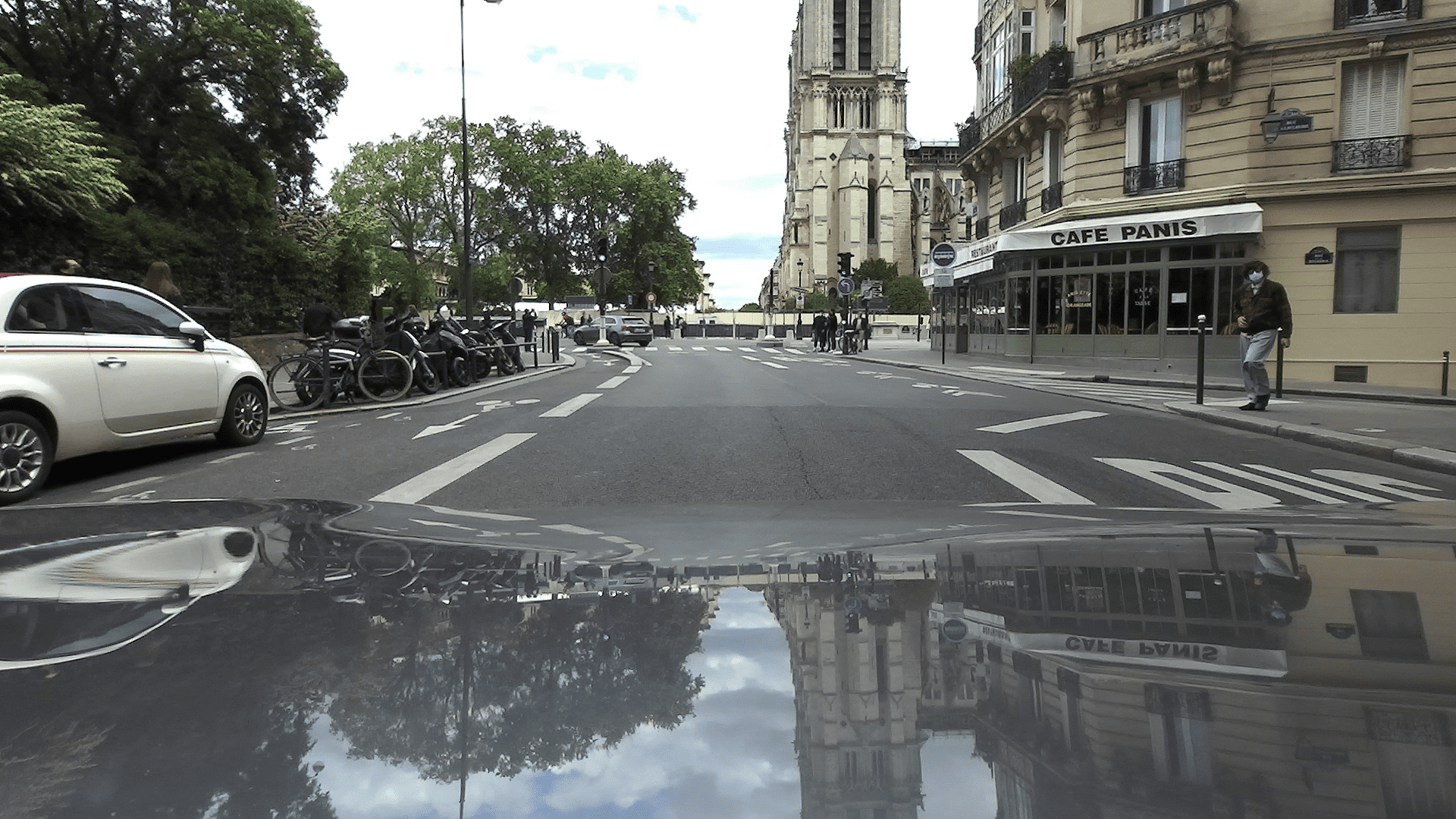}
        \includegraphics[width=\textwidth]{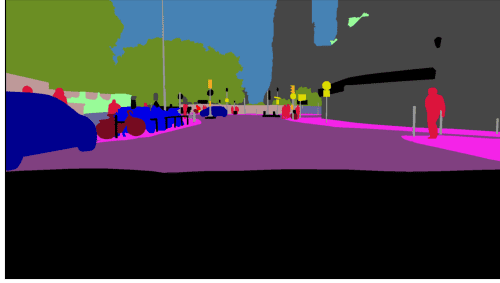}
    \end{subfigure}

    \begin{subfigure}[b]{0.25\textwidth}
        \includegraphics[width=\textwidth]{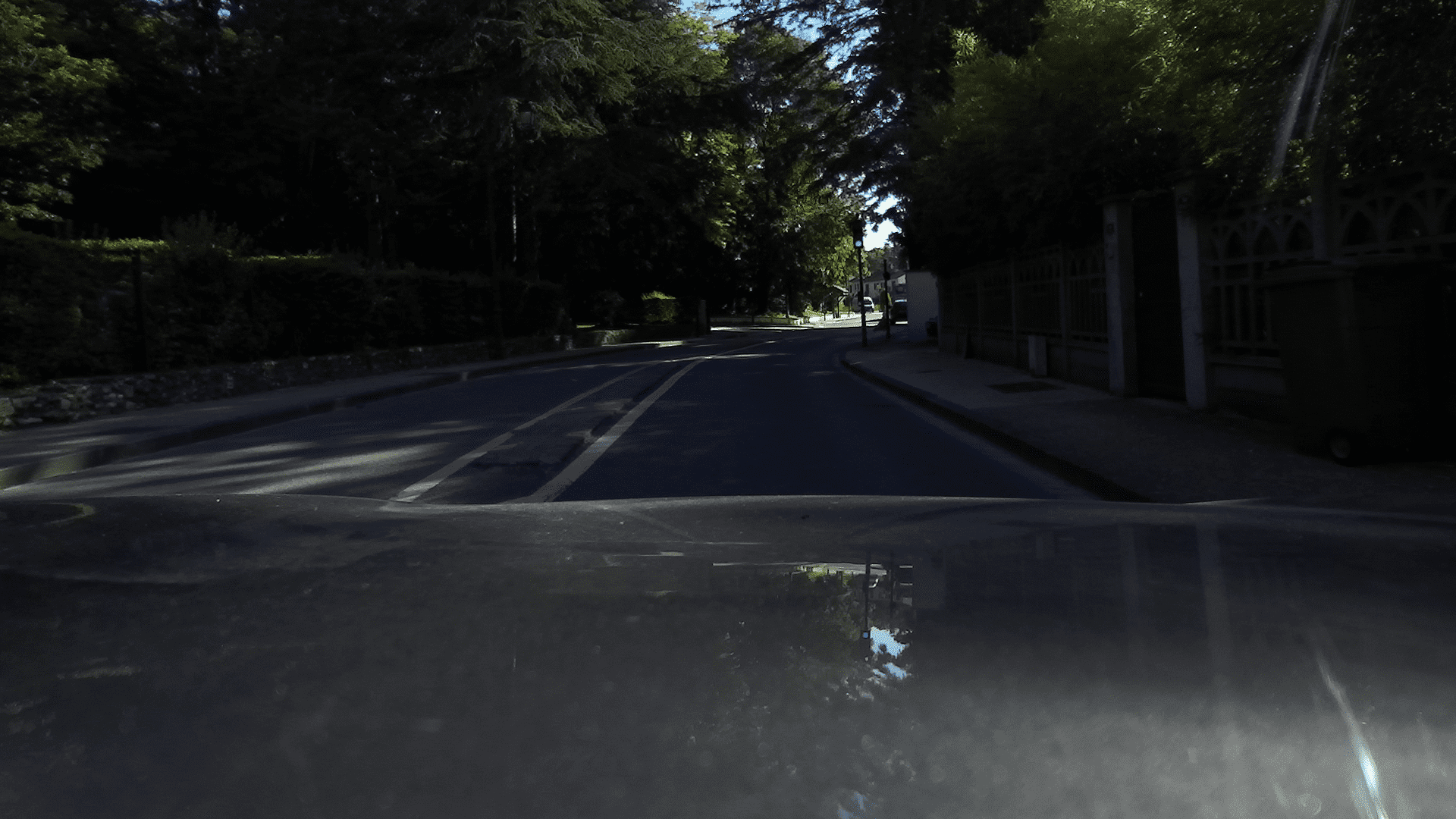}
        \includegraphics[width=\textwidth]{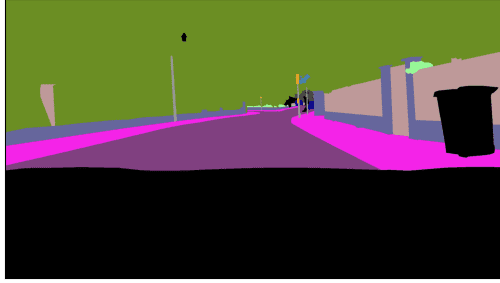}
    \end{subfigure}
    \begin{subfigure}[b]{0.25\textwidth}
        \includegraphics[width=\textwidth]{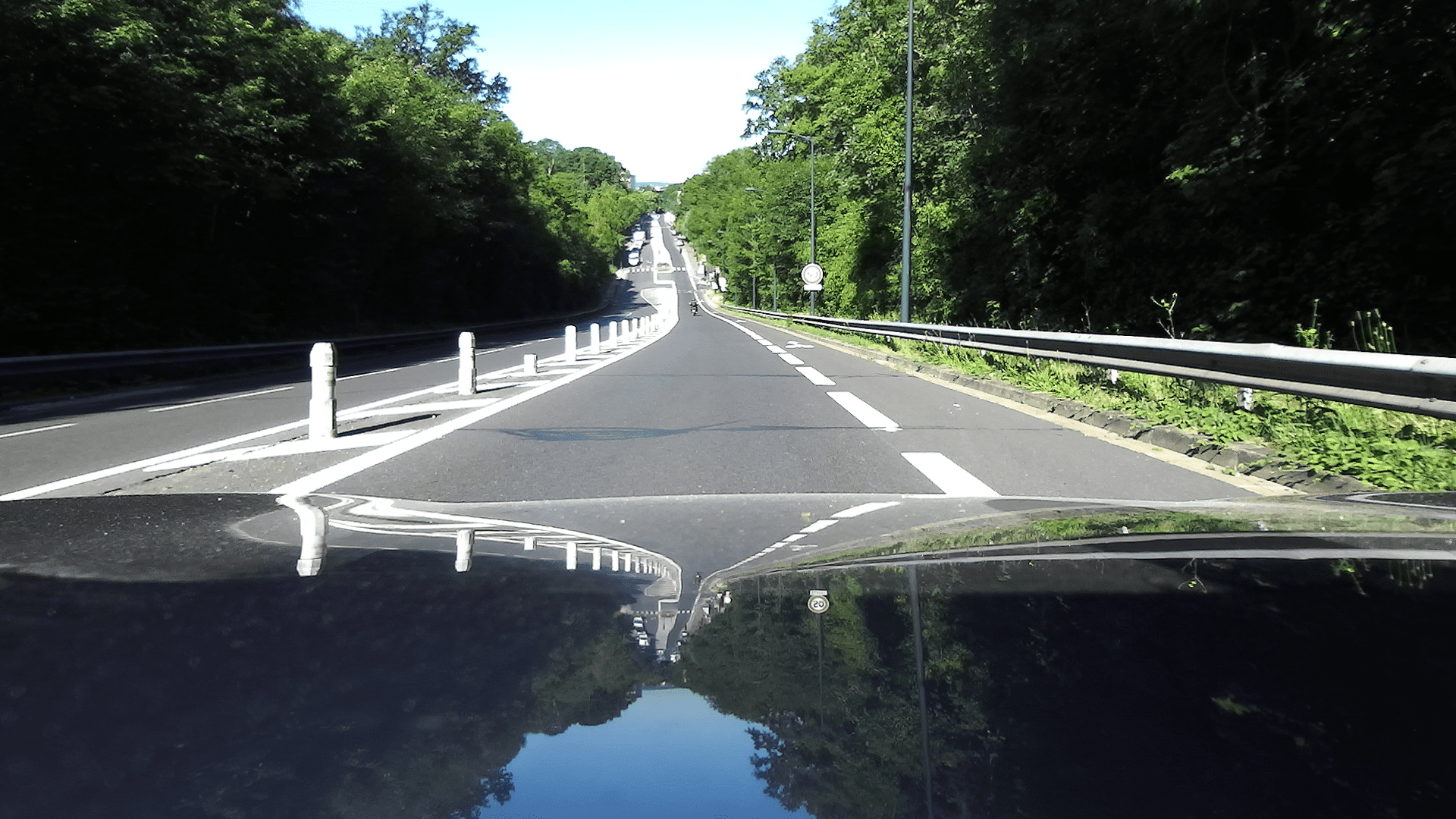}
        \includegraphics[width=\textwidth]{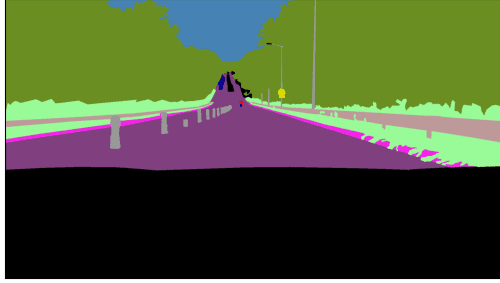}
    \end{subfigure}
    \begin{subfigure}[b]{0.25\textwidth}
        \includegraphics[width=\textwidth]{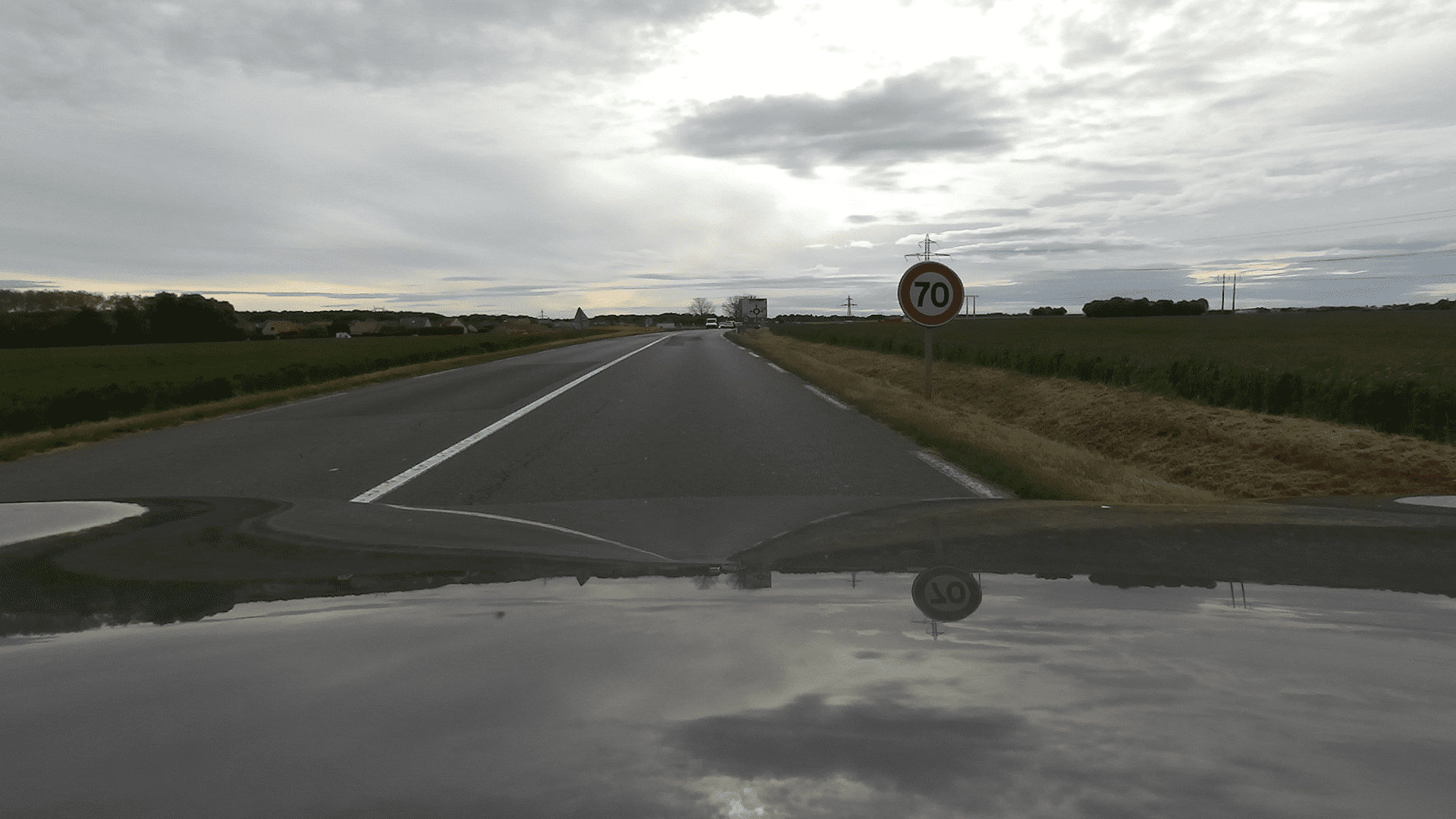}
        \includegraphics[width=\textwidth]{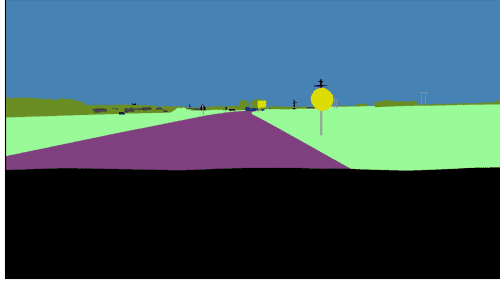}
    \end{subfigure}
    \caption{ Qualitative examples of InfraParis RGB images and their corresponding annotations.}\label{sampleinfraparis1}
\vspace{-1em}
\end{figure*}




\subsection{Camera calibration}\label{subsec:calibration}
\begin{figure}[t!]
    \centering{\includegraphics[width=0.9\linewidth]{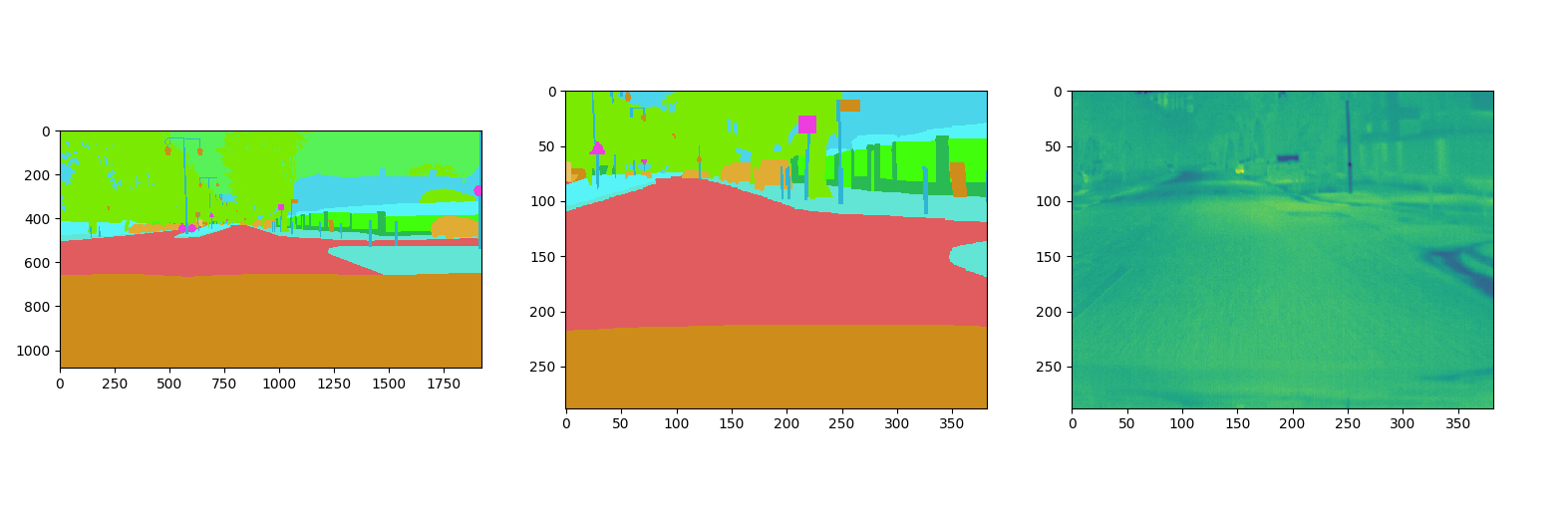}}
    \caption{Results of a re-projection of a semantic map. Left is a semantic image rendered by the ZED. 
    Center is the same image in the 
    \antoine{viewpoint of the Optris.}
    Right is the corresponding 
    \antoine{IR}
    image. Please note the different field of view of the two cameras.}
    \label{fig:calib_image}
    \vspace{-1em}
\end{figure}

Our goal is 
to determine how an image acquired by the ZED camera would look like if it was acquired from the point of view of the IR camera (see Figure \ref{fig:calib_image}) by performing re-projection. In this way any visual task performed on the ZED camera image can be visualized in the IR system coordinate which is critical if we want to combine infrared maps with other modalities. Please note that the field of view of the ZED Camera is larger than the IR one which explains why the IR  is chosen as reference. Indeed, a projection from IR to ZED would have rendered maps with a great number of unknown values.\\

\antoine{The projection matrix, made of the intrinsic parameters of a camera, namely its focal lengths \(\left(f_x, f_y\right)\) and principal point \(\left(c_x, c_y\right)\) is written as follows:}

\begin{small}
\begin{equation} \label{eq:intrinsic_matrix}
\antoine{
\mathcal{K} = 
\begin{pmatrix}
f_x & 0 & c_x\\
0 & f_y & c_y\\
0 & 0 & 1\\
\end{pmatrix}
}
\end{equation}
\end{small}

\begin{figure}[t!]
    \centering{\includegraphics[width=0.8\linewidth]{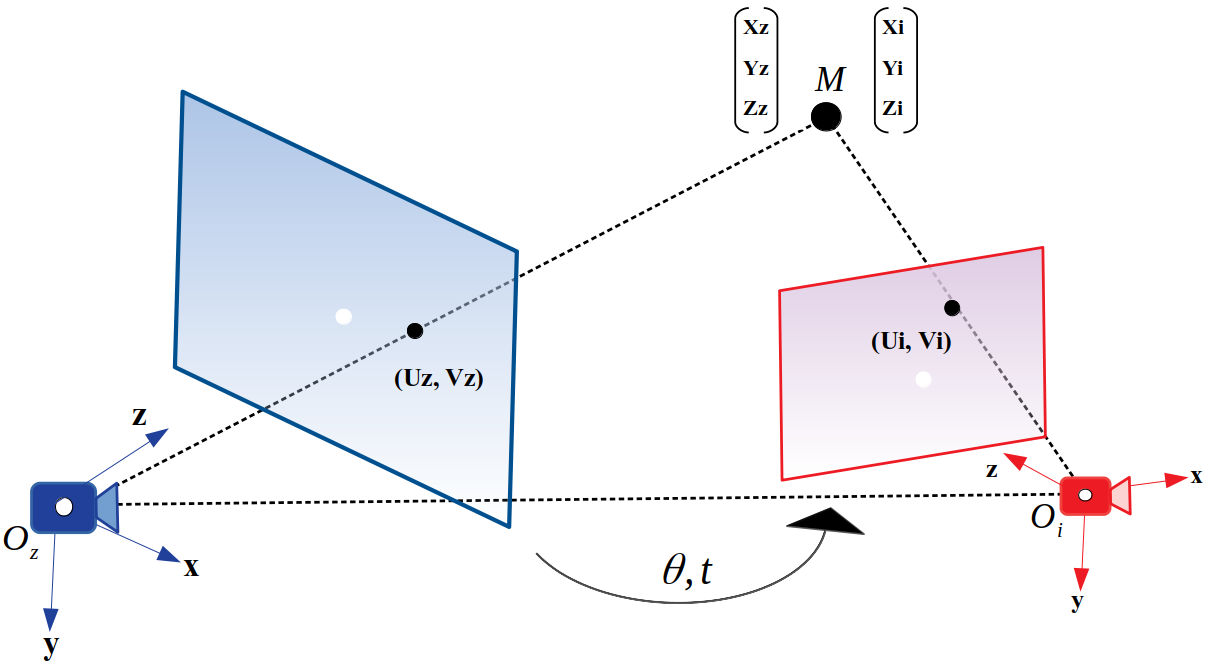}}
    \caption{Our setup consists of a ZED Camera (blue) and an IR Camera (red). The same point \textbf{M} can be written in two different system coordinates. Please note the ZED is composed of two cameras. Here we only displayed the left one as all visual tasks are, by convention, rendered in the left system coordinate.}
    \label{fig:calib_figure}
    \vspace{-1em}
\end{figure}

\antoine{We denote as \(K_{z}\) (resp. \(K_{i}\)) the projective matrix of the ZED (resp. of the Optris IR) camera.}
Let us also define \(\mathcal{R}\) and \(t\) respectively the rotation matrix (illustrated as \(\theta\) in Figure \ref{fig:calib_figure}) and the translation vector of the ZED camera with respect to the 
\antoine{Optris.}
These are the extrinsic parameters,
\antoine{forming the displacement matrix:} 

\begin{small}
\begin{equation}
\mathcal{P}_{Z\rightarrow I} =
\begin{pmatrix}
\mathcal{R} & t
\end{pmatrix}
\end{equation}
\end{small}

\begin{figure}[t!]
    \centering{\includegraphics[width=0.9\linewidth, height=0.15\linewidth]{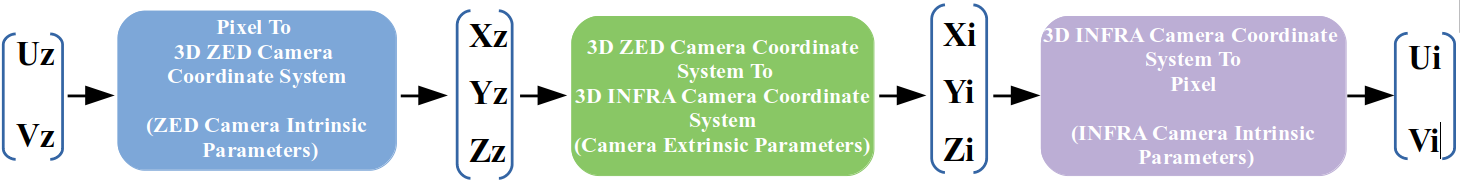}}
    \caption{Pipeline to project a pixel from the ZED 
    \antoine{to the IR}
    Camera.}
    \label{fig:calib_diagram}
    \vspace{-1em}
\end{figure}

Let us consider a specific visual task \(\mathcal{I}\) performed on the ZED camera image, semantic segmentation for instance. And let us consider a given pixel \(\left(U_z, V_z\right)\). The goal of the re-projection is to find the corresponding pixel  \(\left(U_i, V_i\right)\) in the IR system coordinate as described in Figure \ref{fig:calib_figure}. To do so 
\antoine{we apply the following pipeline (see Figure.~\ref{fig:calib_diagram}).}

The 3D point in the ZED camera system coordinate can be obtained from \(\left(U_z, V_z\right)\) as follows:

\begin{small}
\begin{equation}
\begin{pmatrix}
X_z\\
Y_z\\
Z_z\\
\end{pmatrix}
= 
\mathcal{D}\Bigr[U_z, V_z\Bigr]
\antoine{\mathcal{K}_{z}^{-1}}
\begin{pmatrix}
U_z\\
V_z\\
1\\
\end{pmatrix}
\end{equation}
\end{small}

Where \(\mathcal{D}\Bigr[U_z, V_z\Bigr]\) is the depth value at pixel \(\left(U_z, V_z\right)\) as rendered by the ZED camera in the left image coordinate system.

The obtained 3D vector can then be written in the IR system coordinate by using the extrinsic matrix:

\begin{small}
\begin{equation}
\begin{pmatrix}
X_i\\
Y_i\\
Z_i\\
\end{pmatrix}
= 
\mathcal{P}_{Z\rightarrow I}
\begin{pmatrix}
X_z\\
Y_z\\
Z_z\\
1\\
\end{pmatrix}
\end{equation}
\end{small}

\antoine{and}

\begin{small}
\begin{equation}
\begin{pmatrix}
U_i\\
V_i\\
1\\
\end{pmatrix}
= \frac{1}{Z_i}
\antoine{\mathcal{K}_{i}}
\begin{pmatrix}
X_i\\
Y_i\\
Z_i\\
\end{pmatrix}
\end{equation}
\end{small}

\antoine{Finally the}
corresponding re-projected 
\antoine{labelling image $\widetilde{\mathcal{I}}$}
is filled as follows:

\begin{small}
\begin{equation}
    \antoine{\widetilde{\mathcal{I}}}
    \Bigr[U_i, V_i\Bigr] = \mathcal{I}\Bigr[U_z, V_z\Bigr]
\end{equation}
\end{small}

Please note that in order to keep the number of parameters to be determined low we chose not to take into account eventual distortions in our pipeline. We think it was unnecessary considering that images don't seem to be distorted visually. Yet this is something to be considered in the general case.\\

We consider two different kinds of parameters:

\begin{itemize}
    \item \textbf{Known Parameters:} The ZED SDK API provides precise values of the intrinsic parameters of the ZED Camera.

    \item \textbf{Coarse parameter estimates:} The intrinsic parameters of the IR Camera are roughly given by the documentation of the camera
    \antoine{(Optris}
    PI 450i).\\
    For the extrinsic parameters, the IR camera was placed in the middle of the ZED cameras
    Hence, we expect \(t_x\) the first component of the vector \(t\) to be about half the baseline of the ZED Camera. In the same way, \(t_z\) as well as the rotation angle are expected to be around zero.

\end{itemize}

We then refined the coarse parameter estimates
\antoine{by a grid search around}
the coarse parameter values. We chose the parameters that maximize the correlation between the edges of the semantic map and the edges of the infrared image. Edges of the infrared images were obtained with a threshold on the gradient. Edge maps were dilated to improve the metric quality.   

\subsection{Statistics of the dataset}

We captured a set of 12\,084 images in various areas around Paris. During the image capture process, a portion of the acquired images proved to be unusable, ranging from $18\%$ to $60\%$ based on specific acquisitions. These discrepancies arose due to a range of corruptions: some emerged due to sensor malfunctions blending two images (as illustrated in Figure \ref{fig:useless_images}), while others were marred by noise and unconventional lighting. \xuanlong{In Appendix Section B, we provide an overview of instances where annotation ambiguities arose and detail the strategies we employed to effectively resolve them.}

Consequently, we meticulously curated a total of 7\,301 viable images, which we subsequently categorized into three distinct groups: \textit{Train}, \textit{Validation}, and \textit{Test}, as delineated in Table \ref{tab:database_statistics}.  The objective was to assemble a diverse and cohesive dataset with validation and test subsets that amalgamate both rural and urban contexts. \gianni{Nonetheless, the InfraParis dataset offers an additional 
\antoine{set of 16\,142}
consecutive frames, which can be interpreted as video data. These video sequences prove particularly valuable for the unsupervised depth task due to their potential to enhance depth estimation accuracy.}

The distribution of these 7\,301 images is thoughtfully arranged across various cities surrounding the Paris area, as detailed in Table \ref{tab:database_statistics}. 
This geographical spread affords a heterogeneous assortment of scenes. For instance, the core of Paris exhibits a bustling atmosphere with towering modern structures in the 13th arrondissement juxtaposed with Haussmannian buildings in the 5th. Meanwhile, the suburbs like Orsay showcase a more organized layout of quaint houses, while Clamart and Meudon, situated near forests, evoke a rural ambiance. 
Examples of images from the dataset are depicted in Figures \ref{sampleinfraparis1}.

In an effort to inject diversity among the different cities, we chose to employ a subset of images for each city, ensuring an absence of overlap between consecutive frames.

\begin{figure}[t!]
    \centering
    \begin{subfigure}[b]{0.20\textwidth}
        \includegraphics[width=\textwidth]{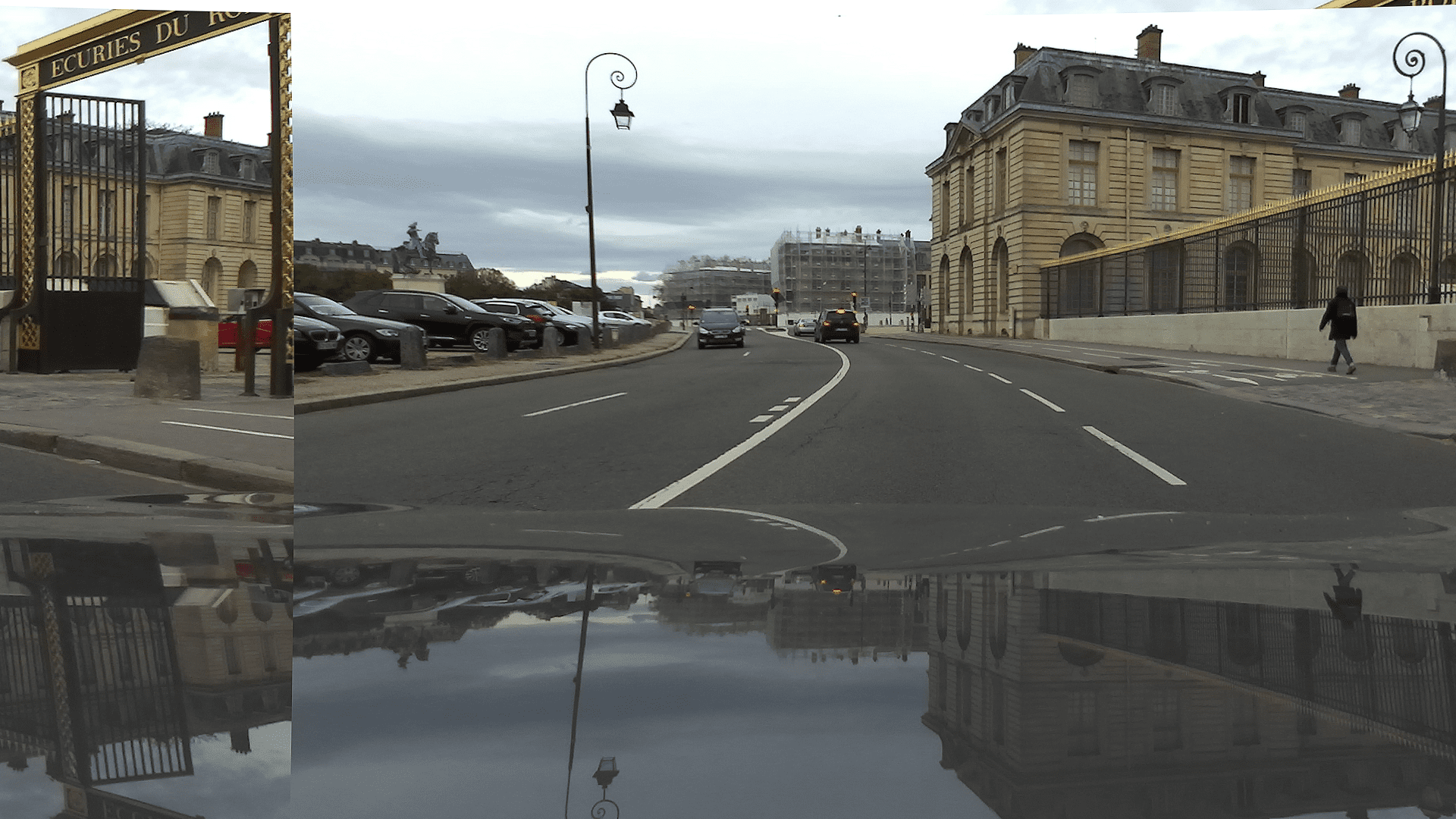}
    \end{subfigure}
    \begin{subfigure}[b]{0.20\textwidth}
        \includegraphics[width=\textwidth]{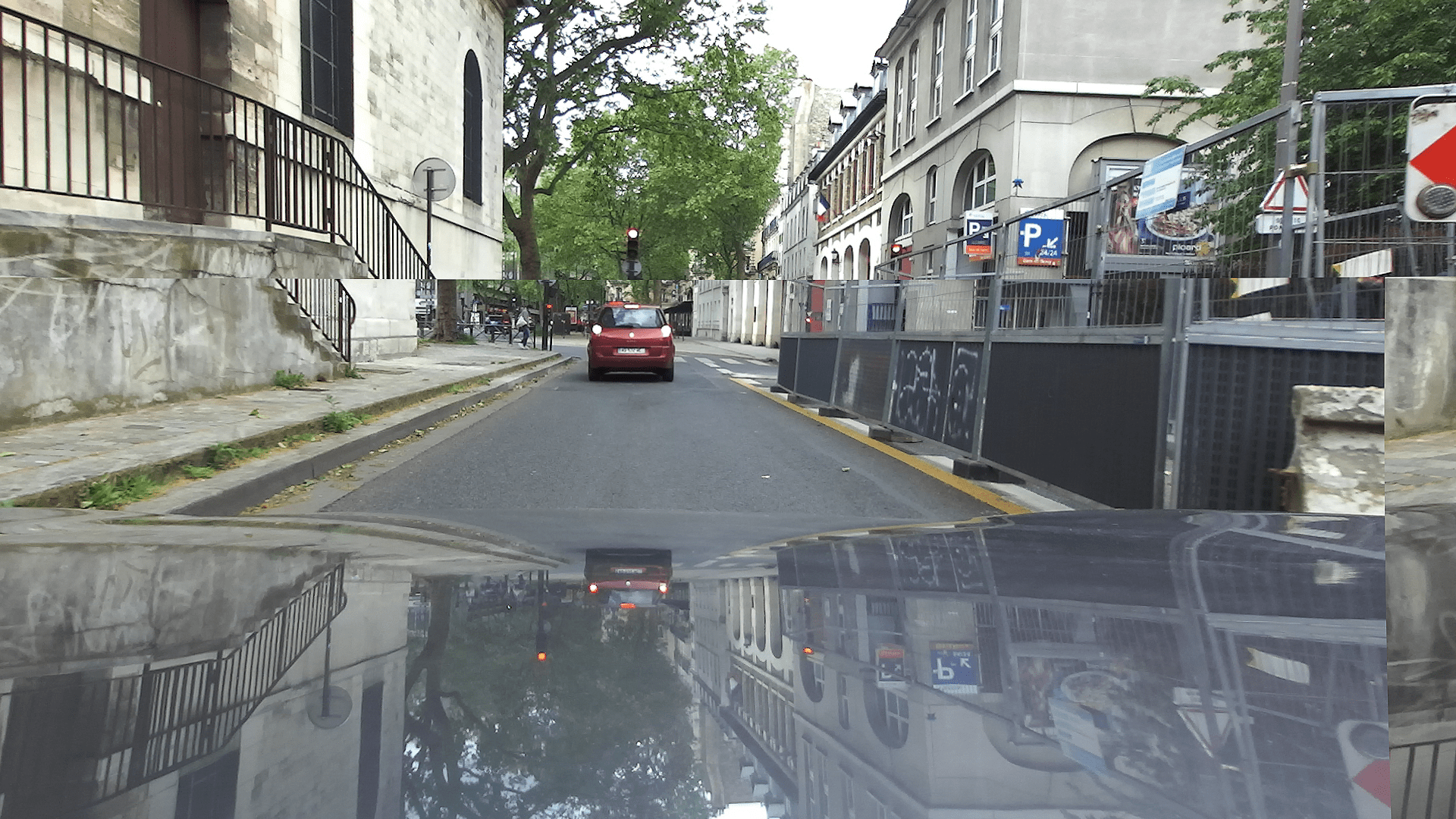}
    \end{subfigure}
    \caption{Example of unusable images from the capture due to registration artifacts.}
    \label{fig:useless_images}
\end{figure}

\begin{table}[!h]
\centering
\scalebox{0.7}{%

\begin{tblr}{
  cell{2}{1} = {r=10}{},
  cell{12}{1} = {r=6}{},
  cell{18}{1} = {r=3}{},
  hline{1-2,21-22} = {-}{0.08em},
  hline{12,18} = {-}{0.05em},
}
 & \textbf{City} & {\textbf{Selected}\\\textbf{images}} & {\textbf{Usable}\\\textbf{images}} & {\textbf{Unusable}\\\textbf{images}} & {\textbf{Percentage}\\\textbf{\textbf{of~}unusable}\\\textbf{images}}\\
\begin{sideways}\textbf{Train}\end{sideways} & Antony BLR & 1035 & 1320 & 710 & 34.98\%\\
 & Bièvres & 661 & 848 & 238 & 21.92\%\\
 & Chaville-Sevre-Viroflay & 906 & 1199 & 332 & 21.69\%\\
 & Meudon-Clamart & 100 & 2076 & 452 & 17.88\%\\
 & Orsay Saclay & 631 & 986 & 276 & 21.87\%\\
 & Paris5-6 & 1105 & 1415 & 2016 & 58.76\%\\
 & SQY & 99 & 283 & 164 & 36.69\%\\
 & SQY-Montigny & 782 & 946 & 319 & 25.22\%\\
 & Trappes & 499 & 1220 & 694 & 36.26\%\\
 & Versailles & 749 & 884 & 539 & 37.88\%\\
\begin{sideways}\textbf{Validation}\end{sideways} & Massy & 40 & 973 & 589 & 37.71\%\\
 & Palaiseau & 10 & 436 & 584 & 57.25\%\\
 & Paris13 & 73 & 114 & 1390 & 92.42\%\\
 & Paris14 & 39 & 64 & 379 & 85.55\%\\
 & Paris15 & 7 & 12 & 90 & 88.24\%\\
 & Plateau & 20 & 438 & 378 & 46.32\%\\
\begin{sideways}\textbf{Test}\end{sideways} & Paris12 & 190 & 921 & 2273 & 71.16\%\\
 & Saint-Cyr & 82 & 1361 & 450 & 24.85\%\\
 & Verrières & 299 & 646 & 211 & 24.62\%\\
 & \textbf{Total} & \textbf{7301} & 16142 & 12084 & 44.28\%
\end{tblr}
}
\caption{\textbf{Database statistics for each of the cities within the dataset.}}
\label{tab:database_statistics}
\end{table}

\subsection{Class labels}

We provide precise pixel-level annotations for 20 classes. 
Our meticulous annotation effort resulted in a set of 7\,301 finely-detailed images, each adorned with layered polygons. These annotations were accomplished in collaboration with a professional annotation company. On average, approximately 1.5 hours were invested in annotating each image, followed by an additional 20 minutes of quality verification conducted by our team. The annotation team was directed to designate instances as "unlabeled" in cases of uncertainty and to document any new, previously-unlabeled instance types encountered during the process, in order to maintain comprehensive records.

Our annotation schema comprises 20 distinct visual classes, systematically organized into eight overarching categories: flat, construction, nature, vehicle, sky, object, human, and void. Notably, the ``road" class encompasses sections of the ground typically traversed by vehicles, including all lanes, directions, and streets, complete with road markings. Furthermore, areas delimited solely by road markings, such as bicycle lanes, roundabout lanes, and parking spaces, are also classified as \xuanlong{``}road". Curbs, however, are excluded from this label.

The ``Sidewalk" class encapsulates ground segments designated for pedestrians or cyclists, demarcated from the road by obstacles like curbs or poles, rather than mere markings. Often elevated relative to the road, sidewalks are typically situated along road sides. This category encompasses pedestrian zones, walkable parts of traffic islands, and features that create separation from the road.

\xuanlong{The ``person" class} includes individuals walking, standing, or sitting on surfaces such as the ground, benches, or chairs. It also incorporates toddlers, people pushing bicycles, or those standing adjacent to bicycles with both legs on one side. Items carried by a person, like backpacks, are part of this class, but objects in contact with the ground, like trolleys, are not included.

The ``rider" class designates a human employing a device to traverse a distance. This category encompasses riders/drivers of bicycles, motorbikes, scooters, skateboards, horses, rollerblades, wheelchairs, road cleaning cars, and open-top cars. Notably, humans within cars are encompassed by the ``car" label, as the label does not account for holes or openings.

The ``car" class encompasses vehicles such as cars, jeeps, SUVs, vans with continuous body shapes, caravans, and excludes other types of trailers. The ``truck" class encompasses trucks, box trucks, and pickup trucks, along with their associated trailers. Notably, the back portion or loading area is physically separated from the driving compartment. The ``bus" class pertains to vehicles designed for the transportation of 9 or more individuals, serving either as public transportation or for long-distance travel.

The ``on rails" class pertains to vehicles operating on tracks, including trams and trains. The ``motorcycle and bicycle" class covers motorbikes, mopeds, and scooters without riders (who are referred to as ``riders", as mentioned above), as well as bicycles without riders.

The ``building" class encompasses structures such as buildings, skyscrapers, houses, bus stop buildings, garages, and carports. Even if a building features glass walls through which visibility is possible, the entire structure is categorized as a building. This class also includes scaffolding affixed to buildings. On the other hand, the ``individual standing wall" class pertains to standalone walls that are not part of a larger building.

Table \ref{tab:annotation_Statistics} summarizes the number of images and pixels corresponding to each class. 
\gianni{Only the class ``person'' is annotated for the object detection task.}

\begin{table}
\centering
\scalebox{0.69}{
\begin{tabular}{lcr}
\toprule
\begin{tabular}[c]{@{}l@{}}\textbf{Classe}\\\textbf{names}\end{tabular} & \multicolumn{1}{l}{\begin{tabular}[c]{@{}l@{}}\textbf{ \# images with}\\\textbf{the annotations}\end{tabular}}& \multicolumn{1}{l}{\begin{tabular}[c]{@{}l@{}}\textbf{percent of pixels with}\\\textbf{the annotations}\end{tabular}} \\
\midrule
Road & 7326 & 9,75 \%  \\
\midrule
Sidewalk & 7102 & 3,38 \% \\
\midrule
Building & 6917 & 12,69 \% \\
\midrule
Wall & 4847 & 1,44 \% \\
\midrule
Fence & 5992 & 2,29 \% \\
\midrule
Pole & 7292 & 0,77 \% \\
\midrule
Traffic light & 3231 & 0,08 \% \\
\midrule
Traffic sign & 5479 & 0,18 \% \\
\midrule
Vegetation & 7080 & 13,04 \% \\
\midrule
Terrain & 6145 & 3,29 \% \\
\midrule
Sky & 7260 & 10,10 \% \\
\midrule
Person & 3755 & 0,16 \% \\
\midrule
Rider & 7174 & 0,03 \%  \\
\midrule
Car & 6906 & 3,42 \% \\
\midrule
Truck & 874 & 0,12 \% \\
\midrule
Bus & 686 & 0,13\% \\
\midrule
Train & 20 & 0,00\% \\
\midrule
Motorcycle & 1669 & 0,10 \% \\
\midrule
Bicycle & 1572 & 0,07\% \\
\midrule
Unlabeled & 7301 & 40,89 \%  \\
\bottomrule
\\
\end{tabular}
}
\caption{Overview of annotated classes}
\vspace{-1em}
\label{tab:annotation_Statistics}
\end{table}

\section{Experimental Results}
\subsection{Semantic segmentation}
\label{sec:semantic_seg}

\gianni{In this section, we present a benchmark for semantic segmentation using reference models, specifically SegFormer \cite{xie2021segformer} and DeepLab v3+ \cite{chen2018encoder}. \xuanlong{In Supplementary materials Section A,} we provide the detailed hyperparameters for these experiments.}

\gianni{The outcomes of segmentation Deep Neural Networks (DNNs) trained on the InfraParis (RGB) dataset and tested on both InfraParis and Cityscapes \cite{Cordts2016Cityscapes} are summarized in Table \ref{tab:semantic_seg}. Additionally, we trained the DNNs on Cityscapes and evaluated them on both Cityscapes and InfraParis. Notably, the two datasets exhibit comparable mean Intersection over Union (mIoU) values, indicating similar behavior.}

\gianni{Furthermore, Table \ref{tab:semantic_seg} illustrates the outcomes of training on the infrared images of InfraParis and testing on the same modality, emphasizing that the results are notably lower. This observation is intriguing as it suggests that infrared images alone might not be sufficient for effective semantic segmentation.}

\begin{table}
\centering
\scalebox{0.66}{
\begin{tblr}{
  row{1} = {c},
  row{2} = {c},
  row{3} = {c},
  cell{1}{1} = {r=3}{},
  cell{1}{2} = {r=3}{},
  cell{1}{3} = {c=2}{},
  cell{4}{1} = {r=8}{c},
  cell{4}{3} = {c},
  cell{4}{4} = {c},
  cell{5}{3} = {c},
  cell{5}{4} = {c},
  cell{6}{3} = {c},
  cell{6}{4} = {c},
  cell{7}{3} = {c},
  cell{7}{4} = {c},
  cell{8}{3} = {c},
  cell{8}{4} = {c},
  cell{9}{3} = {c},
  cell{9}{4} = {c},
  cell{10}{3} = {c},
  cell{10}{4} = {c},
  cell{11}{3} = {c},
  cell{11}{4} = {c},
  cell{12}{1} = {r=8}{c},
  cell{12}{3} = {c},
  cell{12}{4} = {c},
  cell{13}{3} = {c},
  cell{13}{4} = {c},
  cell{14}{3} = {c},
  cell{14}{4} = {c},
  cell{15}{3} = {c},
  cell{15}{4} = {c},
  cell{16}{3} = {c},
  cell{16}{4} = {c},
  cell{17}{3} = {c},
  cell{17}{4} = {c},
  cell{18}{3} = {c},
  cell{18}{4} = {c},
  cell{19}{3} = {c},
  cell{19}{4} = {c},
  cell{20}{1} = {r=8}{c},
  cell{20}{3} = {r=8}{c},
  cell{20}{4} = {c},
  cell{21}{4} = {c},
  cell{22}{4} = {c},
  cell{23}{4} = {c},
  cell{24}{4} = {c},
  cell{25}{4} = {c},
  cell{26}{4} = {c},
  cell{27}{4} = {c},
  hline{1} = {-}{},
  hline{4,28} = {-}{0.08em},
  hline{12,20} = {-}{0.05em},
}
{Training\\set} & Models & mIoU~$\uparrow$ & \\
 &  & Eval set 1 & Eval set 2\\
 &  & Cityscapes & InfraParis\\
\begin{sideways}Cityscapes\end{sideways} & DeepLabV3+MobileNet & 72.767 & 51.926\\
& DeepLabV3+Resnet101 & 77.122 & 55.815\\
 & Segformer B0 & 72.874 & 54.630\\
 & Segformer B1 & 75.068 & 57.229\\
 & Segformer B2 & 78.972 & 59.859\\
 & Segformer B3 & 80.201 & 60.784\\
 & Segformer B4 & 80.008 & 62.463\\
 & Segformer B5 & 80.994 & 62.995\\
\begin{sideways}InfraParis RGB\end{sideways} & DeepLabV3+MobileNet & 47.685 & 65.651\\
& DeepLabV3+Resnet101 & 53.062 & 69.040\\
 & Segformer B0 & 55.051 & 64.160\\
 & Segformer B1 & 58.369 & 68.006\\
 & Segformer B2 & 63.589 & 69.852\\
 & Segformer B3 & 61.775 & 68.803\\
 & Segformer B4 & 63.583 & 70.333\\
 & Segformer B5 & 63.853 & 70.595\\
\begin{sideways}InfraParis Thermal\end{sideways} & DeepLabV3+MobileNet &  & 31.158\\
& DeepLabV3+Resnet101 & - & 34.445\\
 & Segformer B0 &  & 31.032\\
 & Segformer B1 &  & 35.313\\
 & Segformer B2 &  & 35.313\\
 & Segformer B3 &  & 36.623\\
 & Segformer B4 &  & 36.708\\
 & Segformer B5 &  & 36.161
\end{tblr}
}
\caption{\textbf{Comparative results for semantic segmentation task.} \gianni{ It is important to emphasize that within the InfraParis dataset, training and testing occur on the same type of images—whether they are RGB images or thermal infrared images.}}
\label{tab:semantic_seg}
\end{table}

\subsection{Supervised monocular depth estimation}
\label{sec:supervised_mono_depth}

\begin{table*}[t!]
\centering
\scalebox{0.72}{
\begin{tabular}{llccccccc} 
\toprule
Training set & Eval set & Abs Rel~$\downarrow$ & Sqr Rel~$\downarrow$ & RMSE~$\downarrow$ & RMSElog~$\downarrow$ & $\delta<1.25 \uparrow$ & $\delta<1.25^2 \uparrow$ & $\delta<1.25^3 \uparrow$ \\ 
\toprule
\multirow{3}{*}{KITTI} & KITTI & 0.049 & 0.095 & 1.311 & 0.071 & 0.982 & 0.998 & 1.000 \\
 & InfraParis & 0.388 & 4.486 & 10.543 & 0.747 & 0.339 & 0.565 & 0.677 \\
 & Cityscapes & 0.328 & 4.324 & 10.788 & 0.709 & 0.481 & 0.609 & 0.684 \\ 
\midrule
\multirow{3}{*}{\begin{tabular}[c]{@{}l@{}}InfraParis \\RGB\end{tabular}} & KITTI & 0.267 & 1.674 & 4.868 & 0.267 & 0.573 & 0.887 & 0.986 \\
 & InfraParis & 0.203 & 0.860 & 3.638 & 0.234 & 0.680 & 0.945 & 0.987 \\
 & Cityscapes & 0.324 & 2.448 & 7.537 & 0.469 & 0.244 & 0.553 & 0.862 \\ 
\midrule
\begin{tabular}[c]{@{}l@{}}InfraParis\\Thermal\end{tabular} & InfraParis & 0.152 & 0.530 & 2.637 & 0.183 & 0.812 & 0.969 & 0.993 \\
\bottomrule
\end{tabular}
}
\caption{\textbf{Comparative results for supervised monocular depth estimation.} The evaluation depth range is 0-40 meters.}
\label{tab:supervised_depth}
\end{table*}

\begin{table}[t!]
\centering
\scalebox{0.7}{
\begin{tabular}{lcccccc} 
\toprule
Model & AP & AP50 & AP75 & APs & APm & APl \\ 
\toprule
Faster R-CNN & 24.825 & 44.363 & 23.963 & 7.148 & 36.774 & 64.205\\
Mask R-CNN & 29.935 & 52.685 & 26.99 & 10.791 & 41.285 & 65.685\\
\bottomrule
\end{tabular}
}

\caption{\textbf{Comparative results for object detection.} Models were pretrained on COCO and then finetuned on InfraParis. The threshold score for the region of interests was set to 0.7.}
\label{tab:object_detection}
\end{table}

We here provide a benchmark for supervised monocular depth estimation. The baseline model is established using NeWCRFs~\cite{yuan2022newcrfs}, which employs a Swin-Transformer~\cite{liu2021Swin} as the encoder.  We conduct the following experiments to provide the benchmark as well as show the 
\antoine{versatility}
of the proposed dataset.

In the first experiment, the model is trained on the InfraParis training and validation set, then evaluated on the InfraParis test set, Cityscapes validation set~\cite{Cordts2016Cityscapes}, and KITTI~\cite{geiger2013vision} eigen-spilt~\cite{eigen2014depth} validation set, respectively. The second experiment was performed in the opposite way, training the model on the KITTI dataset and evaluating the performance on the InfraParis test set and Cityscapes. Note that since the depth value range acquired by the sensor is 0-40 m, our evaluation on the KITTI and Cityscapes datasets follows the same range for a reasonable comparison. 
In the third experiment, we train the NeWCRFs model to fit thermal images to the depth values for the corresponding areas.
All training settings are the same as those used for NeWCRFs training on KITTI, except that we use 4 as the batch size on InfraParis RGB images training. The evaluation metrics
\antoine{follow those commonly used in depth maps prediction literature}
~\cite{eigen2014depth,yuan2022newcrfs}.

The benchmark is presented in Table~\ref{tab:supervised_depth}. We can see that the model trained on KITTI cannot directly transfer well to the Cityscapes and InfraParis. Yet, the one trained on InfraParis can provide better performance when it is directly evaluated on the other datasets. Since monocular depth estimation is an ill-posed problem and heavily depends on the training dataset, this benchmark shows the good diversity of the proposed dataset. The model trained on InfraParis thermal images shows even better results. Since the resolution of the thermal data is smaller, we consider that the scene has lower diversity in this case, and one cannot directly compare this result to the previous ones. We take this result as a benchmark of the thermal-to-depth estimation task of the proposed dataset.


\subsection{Object detection}
\label{sec:obj_detection}

In this section, we present a benchmark for object detection using Faster R-CNN~\cite{ren2015faster} and Mask R-CNN~\cite{he2017mask} \xuanlong{architectures with ResNet50~\cite{he2016deep} as the backbone}. Supplementary materials Section A provides detailed hyperparameters. The goal is to detect the class Person. While only one class is considered in our study, it remains a challenging issue \xuanlong{to take} into account the \xuanlong{small} number of samples available for finetuning (3721 useful images with at least one person annotated). We used the library Detectron2~\xuanlong{\cite{wu2019detectron2}} to do our experiments. Models were initialized with the available pre-trained weights on COCO~\xuanlong{\cite{lin2014microsoft}}, then finetuned on the InfraParis training set, and finally evaluated on the InfraParis test set. Results are summarized in Table \ref{tab:object_detection}.

\section{Conclusion}

\gianni{In conclusion, the InfraParis dataset presented in this paper stands as a significant contribution to the field of autonomous driving research. Notably, it introduces a novel multi-modal and multi-task dataset that comprises a total of 7\,301 meticulously annotated multimodal pieces of data. One of the key distinguishing features of this dataset lies in its uniqueness; it is one of the few datasets available that encompasses both multiple tasks and modalities on such a substantial scale.} \gianni{The dataset's value is further amplified by its potential to be seamlessly integrated with existing standard autonomous driving datasets such as Cityscapes or KITTI. By offering an extensive range of data spanning multiple tasks, including semantic segmentation, object detection, and depth prediction, as well as modalities like RGB and infrared, the InfraParis dataset enables comprehensive testing and validation of multi-modal models.}

\gianni{The diverse and challenging scenarios encapsulated within the dataset also make it particularly compelling. The convergence of construction activities related to the Olympic Games and the dynamics introduced by the COVID-19 pandemic have generated a unique amalgamation of scenes that are typically absent from traditional datasets. The dataset thereby empowers researchers to explore new and unconventional scenarios, shedding light on previously unexplored sources of uncertainty and variability.}

\gianni{In essence, the InfraParis dataset bears witness to the progress made in improving the capabilities of autonomous driving systems. By offering a substantial and versatile collection of 
\antoine{multi-modal data with}
multi-task annotations, the dataset not only enhances the training and evaluation of contemporary models but also paves the way for innovations that can contribute to safer and more reliable autonomous vehicles in the future.
}

\textbf{Acknowledgment :}
This work was performed using HPC resources from GENCI-IDRIS (Grant 2021 - AD011011970R1) and (Grant 2022 - AD011011970R2). We gratefully acknowledge the support AID Project ACoCaTherm which supported the creation of the dataset.
\clearpage
{
\bibliographystyle{ieee_fullname}
\bibliography{egbib}
}
\clearpage
\input{supplementary}

\end{document}

%% file: abbrev.tex
\definecolor{dgreen}{rgb}{0.0, 0.5, 0.0}
\definecolor{better}{rgb}{0.19, 0.55, 0.91}
\definecolor{worse}{rgb}{0.82, 0.1, 0.26}


%% file: supplementary.tex
\begin{center}
{\textbf{\large{InfraParis: A multi-modal and multi-task autonomous driving dataset\\}}}
\vspace{0.5em}
\textbf{\large{------ Supplementary Material ------}}
\end{center}

\appendix

\renewcommand{\theequation}{A\arabic{equation}}
\renewcommand{\thetable}{A\arabic{table}}
\renewcommand{\thefigure}{A\arabic{figure}}


\section{Implementation details}

\subsection{Semantic segmentation}

Table~\ref{table:tab3} furnishes comprehensive insights into the hyperparameters and elaborate implementation details pertaining to the semantic segmentation task. In the case of RGB images, our approach incorporates classical data augmentation techniques such as random crop, random color jitters, and random horizontal flip, alongside normalization transformations that involve mean subtraction and division by the standard variation. However, for infrared images, a similar augmentation process is applied, albeit without the inclusion of random color jitters.

\begin{table*}[h]
\begin{center}
\scalebox{0.85}
{
\begin{tabular}{l|c|c|c}
\toprule
Architecture         &Deeplab v3+ & Segformer B0,B1,B2 & Segformer B3,B4,B5  \\ 
\midrule
backbone         &ResNet101 and mobilenet & NA & NA  \\ 
\midrule
output stride       &8 & NA &NA\\ 
\midrule
learning rate         &0.1 &0.1 & 0.0001 \\ 
 \midrule
batch size        &16 & 8 & 3  \\ 
 \midrule
number of train iterations  & 100 & 60 & 60  \\ 
\midrule
 weight decay  &0.0001 & 0.01&  0.01 \\ 
 \midrule
 Optimizer        & SGD & AdamW  & AdamW \\ 
  \midrule
 random crop of training images    & 768 & 768 & 1024 \\ 
\bottomrule
\end{tabular}
}
\end{center}
\caption{
{\textbf{Hyper-parameter configuration used in the semantic segmentation experiments.}}
}\label{table:tab3}
\end{table*}

\subsection{Supervised monocular depth estimation}
As we mentioned in the main paper \S4.2, the hyperparameters we used are the same as the official ones applied on the KITTI dataset, except that we use 4 instead of 8 as the batch size when we train the NeWCRFs model on the InfraParis RGB images. We find that batch size 4 works slightly better than 8 during evaluation.

Additionally, we found that sometimes the radar produced some depth information on parts of the front of the car. Thus, during training on InfraParis, we also crop the training images from the top down to a position of $1.8*352$.

\subsection{Object detection}
Table~\ref{table:object_detection_tab} furnishes comprehensive insights into the hyperparameters and elaborate implementation details regarding the object detection task. 
In the case of RGB images, our approach incorporates classical data augmentation techniques such as random crop, random color jitters, and random horizontal flip, alongside normalization transformations that involve mean subtraction and division by the standard variation. Please refer to the detectron2 library for more specific details on the architecture name mentioned in Table~\ref{table:object_detection_tab}.

\begin{table*}[h]
\begin{center}
\scalebox{0.85}
{
\begin{tabular}{l|c|c}
\toprule

Architecture         & faster\_rcnn\_R\_50\_C4\_1x & mask\_rcnn\_R\_50\_FPN\_3x  \\ 
\midrule
backbone         &ResNet50  & ResnNt50  \\ 
\midrule
learning rate         &0.001 &0.001 \\ 
 \midrule
roi.heads batch size        &256 & 256  \\ 
 \midrule
 batch size        &8 & 8  \\ 
 \midrule
number of train iterations  & 50 & 50\\ 
\midrule
 weight decay  &0.0001 & 0.0001 \\ 
 \midrule
 Optimizer        & SGD & SGD \\ 
\bottomrule
\end{tabular}
}
\end{center}
\caption{
{\textbf{Hyper-parameter configuration used in the object detection experiments.}}
}\label{table:object_detection_tab}
\end{table*}

\subsection{Unsupervised monocular depth estimation}
We here provide a benchmark for unsupervised monocular depth estimation.
We started from Monodepth2 pre-trained on KITTI and then fine-tuned it on the InfraParis training and validation set. Then we evaluated it on the InfraParis test set. Results are shown in Table~\ref{tab:unsupervised_depth}. We followed the same evaluation protocol as Monodepth2 to solve the scale ambiguity. We applied a mask adapted to the InfraParis dataset which gets rid of pixels corresponding to both the front of the car and to the sky.

\section{Resolving ambiguous images}
\label{sec:unusable}

Throughout the annotation process, we encountered various instances of ambiguity. For a visual representation of these complexities, please consult Figures~\ref{fig:supp_annotation}. To ensure the integrity of our dataset, particularly in instances where uncertainty prevailed, we collaborated with the annotation company to categorize dubious instances as "unlabeled." This approach was adopted to prevent any adverse impact on the performance of DNNs stemming from incorrect annotations.

\begin{figure*}[t]
    \centering
    \begin{subfigure}[b]{0.45\textwidth}
        \includegraphics[width=\textwidth]{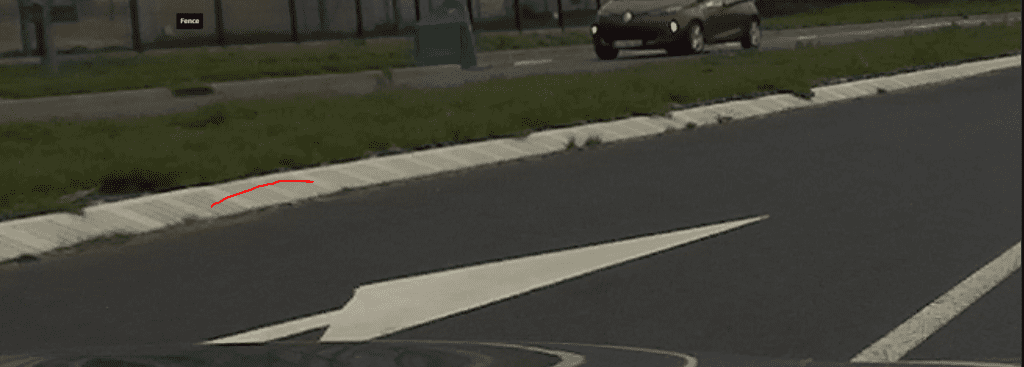}
        \caption{The curb is annotated as the sidewalk.}
        \includegraphics[width=\textwidth]{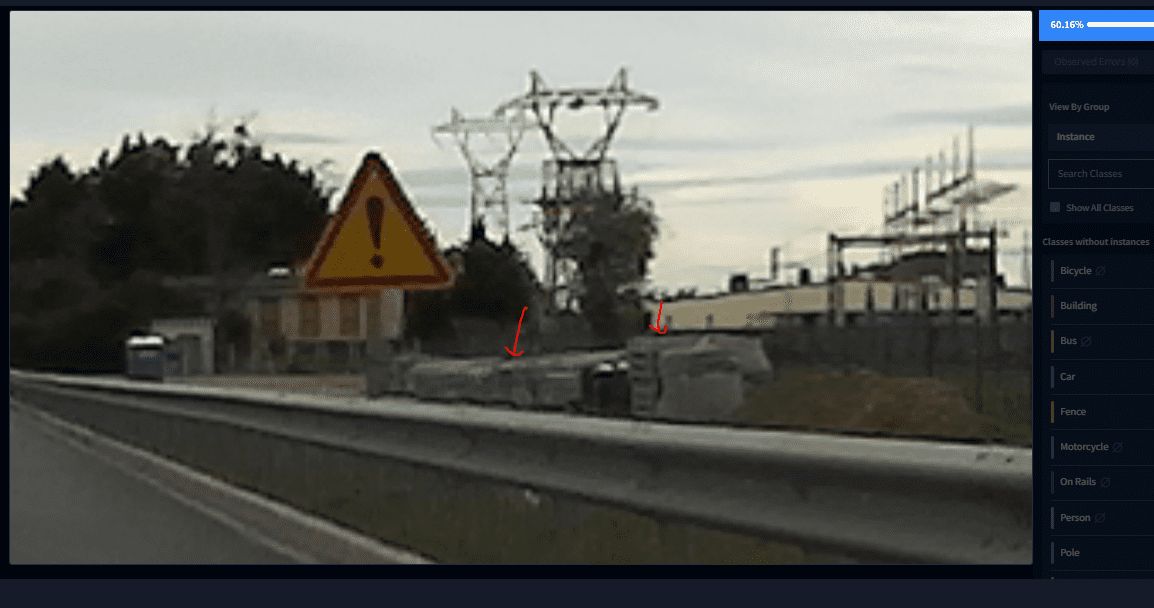}
        \caption{The building materials are annotated as Unlabeled.}
    \end{subfigure}
    \begin{subfigure}[b]{0.45\textwidth}
        \includegraphics[width=\textwidth]{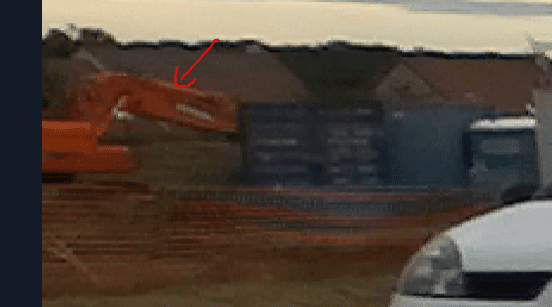}
        \caption{Excavator at the construction site is annotated as Unlabeled.}
        \includegraphics[width=\textwidth]{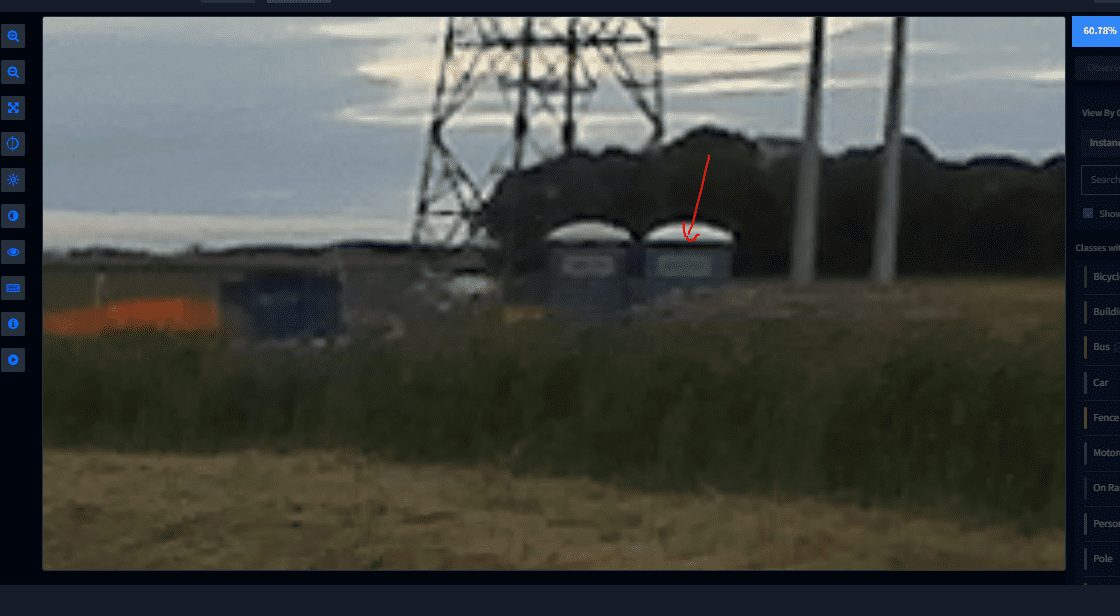}
        \caption{Unknown facilities in the fields are annotated as Unlabeled.}
    \end{subfigure}
    \begin{subfigure}[b]{0.45\textwidth}
        \includegraphics[width=\textwidth]{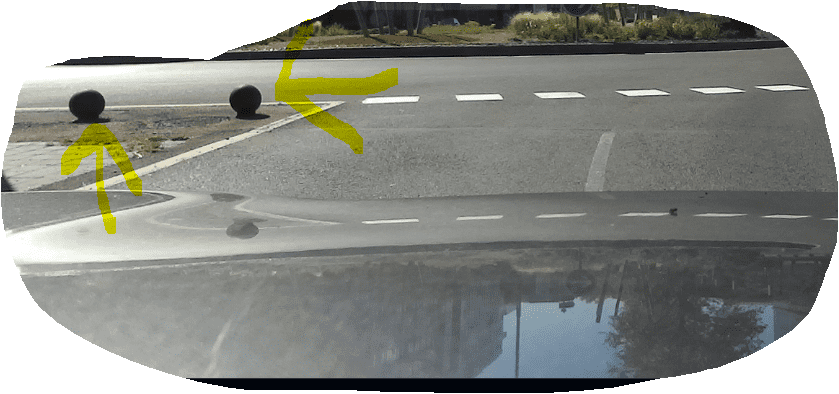}
        \caption{Stone balls on the roadside are annotated as Pole.}
        \includegraphics[width=\textwidth]{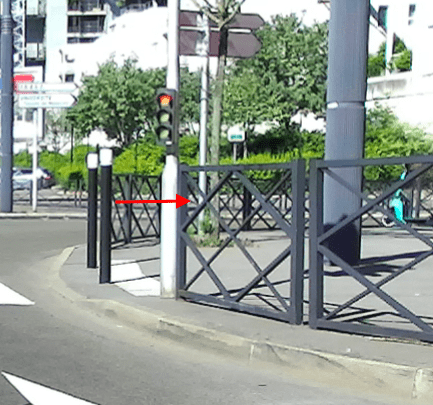}
        \caption{Featured fences are annotated as Fence.}
    \end{subfigure}
    \begin{subfigure}[b]{0.45\textwidth}
        \includegraphics[width=\textwidth]{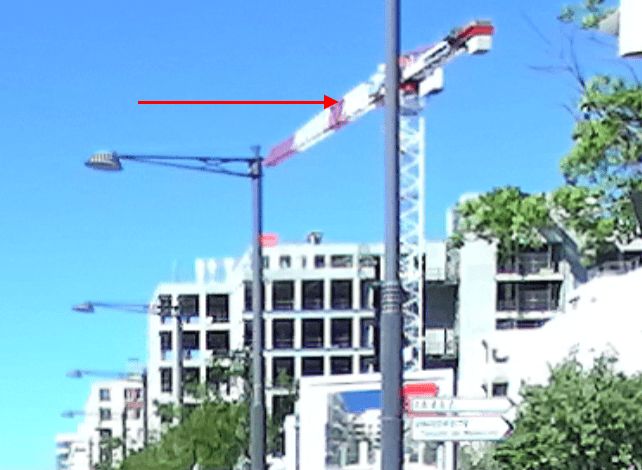}
        \caption{The crane is annotated as unlabeled.}
        \includegraphics[width=\textwidth]{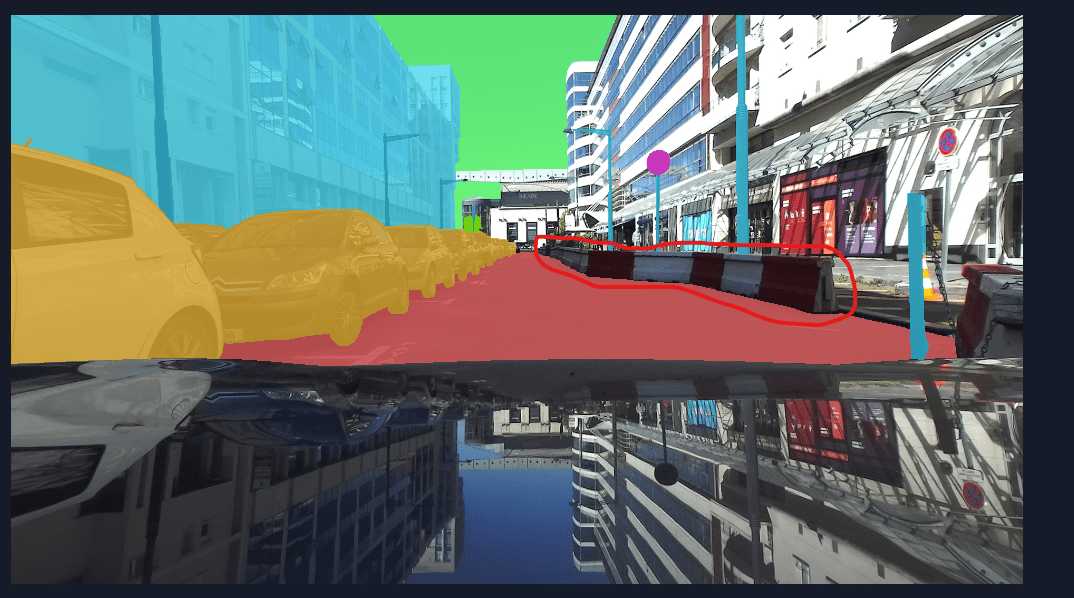}
        \caption{Blocking device at the construction site is annotated as Fence.}
    \end{subfigure}
\end{figure*}

\begin{figure*}[t]\ContinuedFloat
    \centering
    \begin{subfigure}[b]{0.45\textwidth}
        \includegraphics[width=\textwidth]{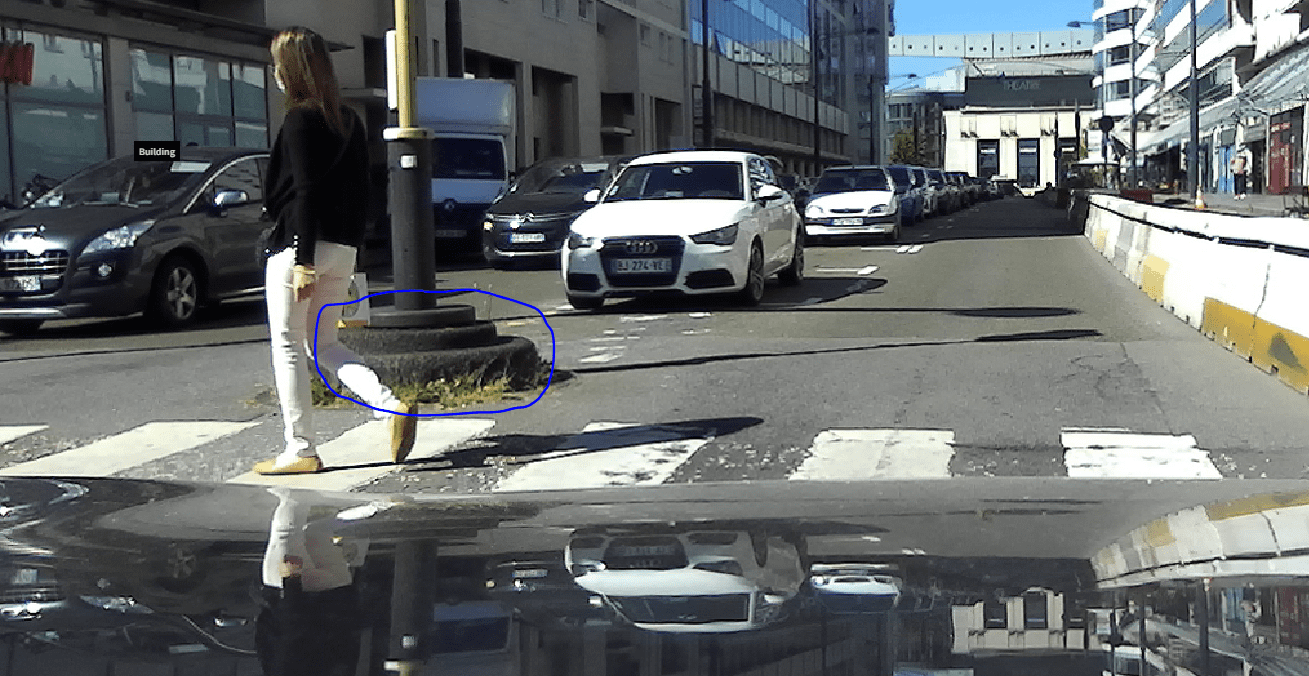}
        \caption{The traffic light base is annotated as Unlabeled.}
        \includegraphics[width=\textwidth]{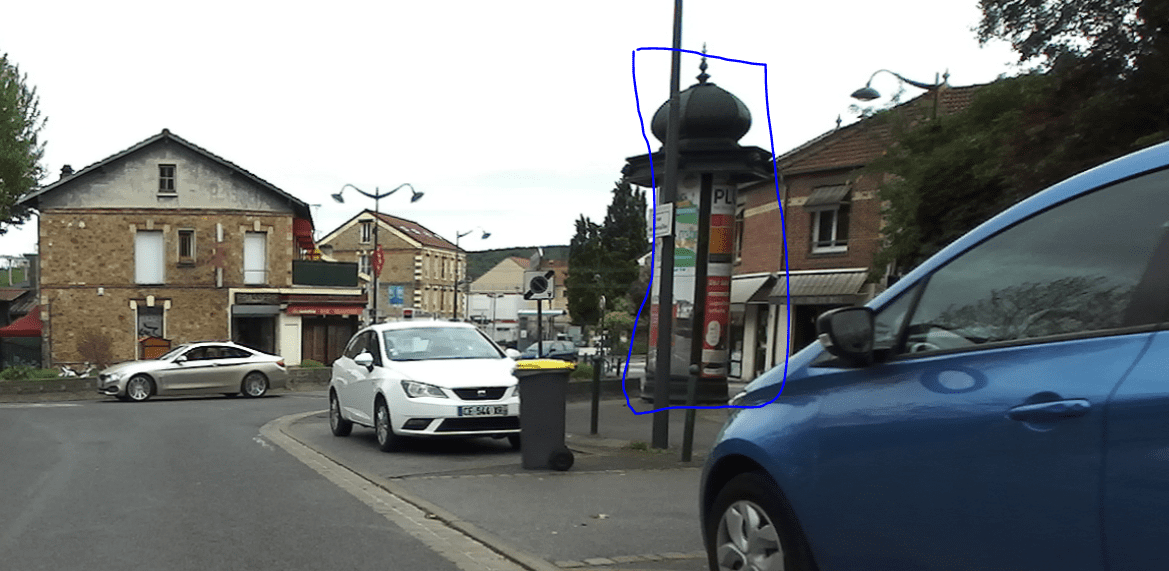}
        \caption{The advertising device is annotated as Unlabeled.}
    \end{subfigure}
    \begin{subfigure}[b]{0.45\textwidth}
        \includegraphics[width=\textwidth]{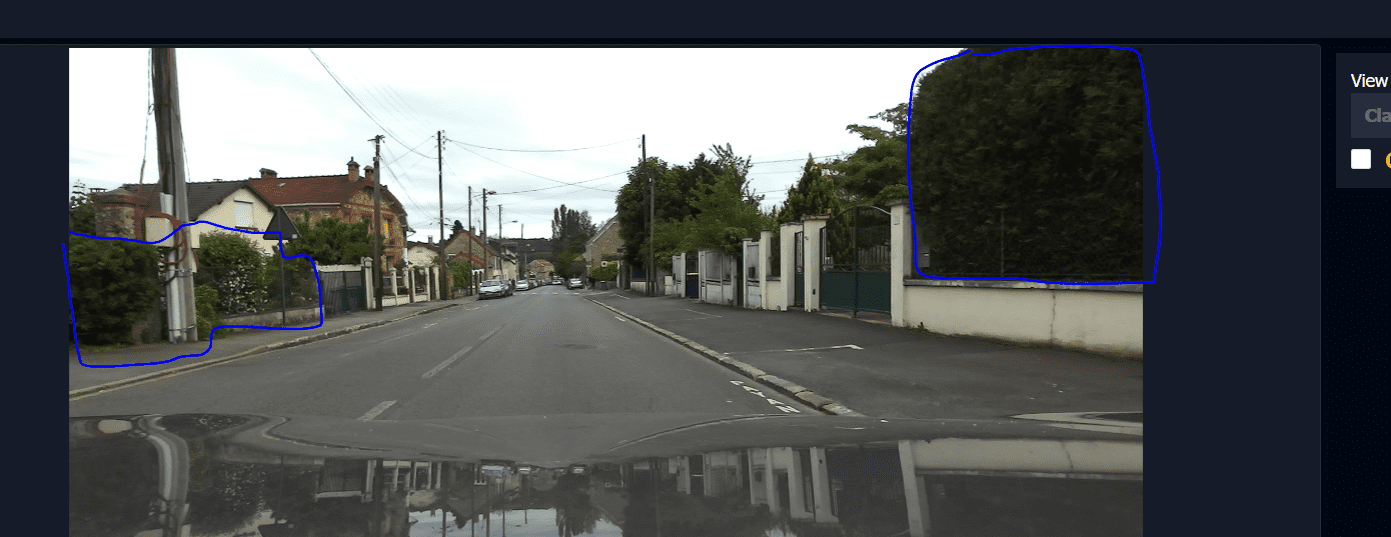}
        \caption{Plants next to the house are annotated as Vegetation.}
        \includegraphics[width=\textwidth]{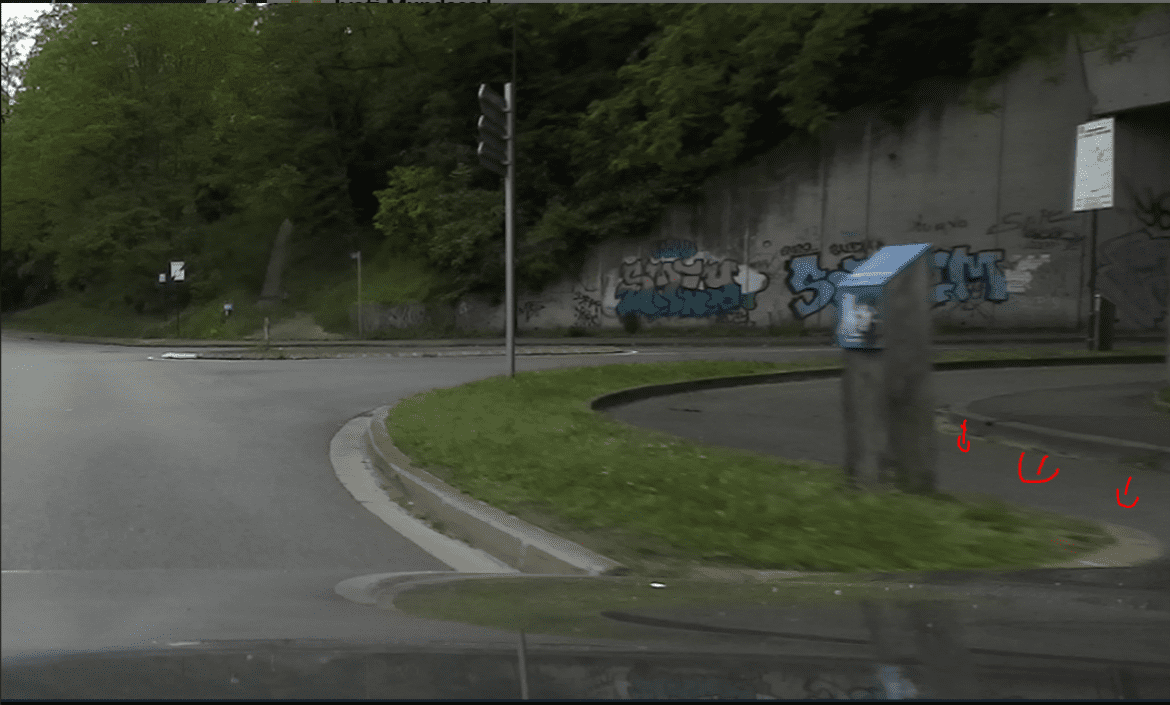}
        \caption{Separation zone between motor vehicle and non-motor vehicle lanes is annotated as Terrain.}
    \end{subfigure}
    \begin{subfigure}[b]{0.45\textwidth}
        \includegraphics[width=\textwidth]{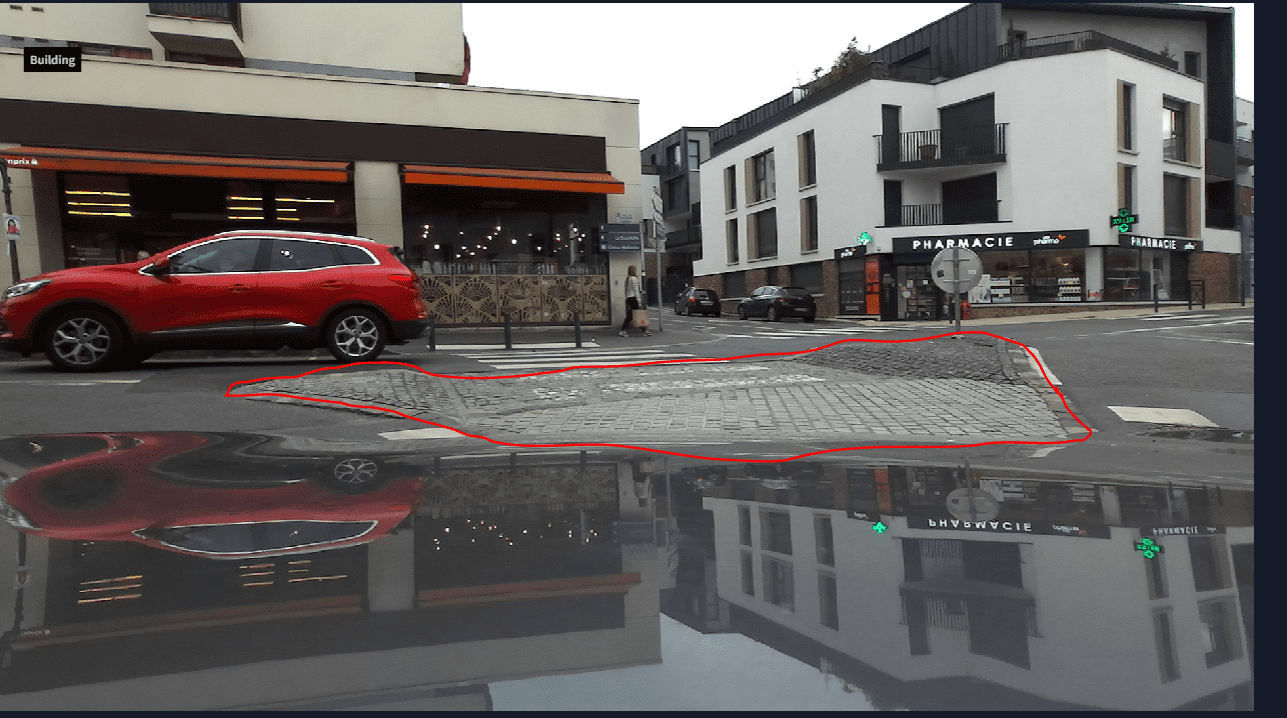}
        \caption{Pedestrian waiting area in the middle of the road is annotated as Side walker.}
        \includegraphics[width=\textwidth]{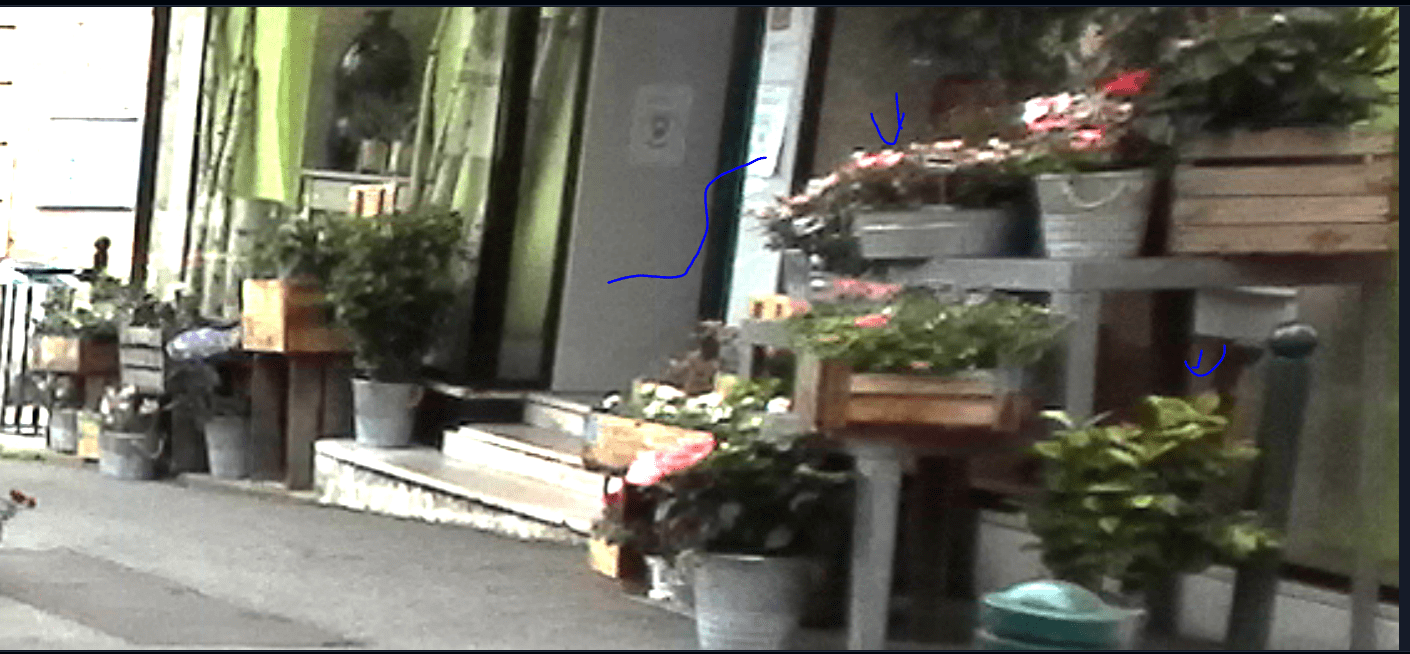}
        \caption{Flowers and pots at the flower shop are annotated as Unlabeled.}
    \end{subfigure}
    \caption{\textbf{Examples of the ambiguous objects encountered in the annotation process.}}\label{fig:supp_annotation}
\end{figure*}

\begin{table*}[t!]
\centering
\scalebox{0.75}{
\begin{tabular}{llccccccc} 
\toprule
Training set & Eval set & Abs Rel~$\downarrow$ & Sqr Rel~$\downarrow$ & RMSE~$\downarrow$ & RMSElog~$\downarrow$ & $\delta<1.25 \uparrow$ & $\delta<1.25^2 \uparrow$ & $\delta<1.25^3 \uparrow$ \\ 
\toprule
\multirow{1}{*}{KITTI} & InfraParis & 0.236 & 0.984 & 3.870 & 0.341 & 0.573 & 0.832 & 0.933 \\
\bottomrule
\end{tabular}
}
\caption{\textbf{Comparative results for unsupervised monocular depth estimation.} The evaluation depth range is 0-40 meters. The model chosen is Monodepth2.}
\label{tab:unsupervised_depth}
\end{table*}